\documentclass[journal]{IEEEtran}
\usepackage{cite}
\ifCLASSINFOpdf

\else

\fi

\usepackage{amsfonts,amssymb}
\usepackage{graphicx}
\usepackage{algorithm}
\usepackage{algorithmic}
\usepackage{amsmath}
\usepackage{booktabs}
\usepackage{multirow}
\usepackage{caption}
\usepackage{color}
\usepackage[switch]{lineno}
\usepackage{ragged2e}
\usepackage{paralist}
\usepackage{enumitem}

\usepackage{times}
\usepackage{amsmath}
\usepackage{amssymb}

\usepackage{subfigure}
\usepackage{cite}

\usepackage{url}
\usepackage{amsfonts,amssymb}
\usepackage{graphicx}
\usepackage{algorithm}
\usepackage{algorithmic}
\usepackage{amsmath}
\usepackage{booktabs}
\usepackage{multirow}
\usepackage{caption}
\usepackage{color}
\usepackage[switch]{lineno}
\usepackage{ragged2e}
\usepackage{paralist}
\usepackage{enumitem}
\usepackage{lineno}

\newtheorem{theorem}{Theorem}

\newtheorem{remark}{Remark}

\begin{document}
% \linenumbers

%\linenumbers

\title{Joint Learning of Neural Transfer and Architecture Adaptation for Image Recognition}

\author{Guangrun Wang, Liang Lin, Rongcong Chen, Guangcong Wang, and Jiqi Zhang
\IEEEcompsocitemizethanks {\IEEEcompsocthanksitem
G. Wang, L. Lin, R. Chen, G. Wang, J. Zhang are with the School of Computer Science and Engineering, Sun Yat-sen University, Guangzhou, P. R. China. L. Lin is also with DarkMatter AI Research, China. Email: wanggrun@mail2.sysu.edu.cn.; linliang@ieee.org; Corresponding author: Liang Lin.}}% <-this % stops a space

\markboth{IEEE TRANSACTIONS ON Neural Networks and Learning Systems}
{G. Wang \MakeLowercase{\textit{et al.}}: Joint Learning of Neural Transfer and Architecture Adaptation for Image Recognition}

\IEEEcompsoctitleabstractindextext{
\begin{abstract}
Current state-of-the-art visual recognition systems usually rely on the following pipeline: (a) pretraining a neural network on a large-scale dataset (e.g., ImageNet) and (b) finetuning the network weights on a smaller, task-specific dataset. Such a pipeline assumes the sole weight adaptation is able to transfer the network capability from one domain to another domain, based on a strong assumption that a fixed architecture is appropriate for all domains. However, each domain with a distinct recognition target may need different levels/paths of feature hierarchy, where some neurons may become redundant, and some others are re-activated to form new network structures. In this work, we prove that dynamically adapting network architectures tailored for each domain task along with weight finetuning benefits in both efficiency and effectiveness, compared to the existing image recognition pipeline that only tunes the weights regardless of the architecture. Our method can be easily generalized to an unsupervised paradigm by replacing supernet training with self-supervised learning in the source domain tasks and performing linear evaluation in the downstream tasks. This further improves the search efficiency of our method. Moreover, we also provide principled and empirical analysis to explain why our approach works by investigating the ineffectiveness of existing neural architecture search. We find that preserving the joint distribution of the network architecture and weights is of importance. This analysis not only benefits image recognition but also provides insights for crafting neural networks. Experiments on five representative image recognition tasks such as person re-identification, age estimation, gender recognition, image classification, and unsupervised domain adaptation demonstrate the effectiveness of our method.
\end{abstract}

% Note that keywords are not normally used for peer-review papers.
}

\maketitle

\IEEEdisplaynotcompsoctitleabstractindextext

\IEEEpeerreviewmaketitle

\section{Introduction}
\label{sect:intro}

\begin{figure*}
\centering
\includegraphics[width=1.0\textwidth]{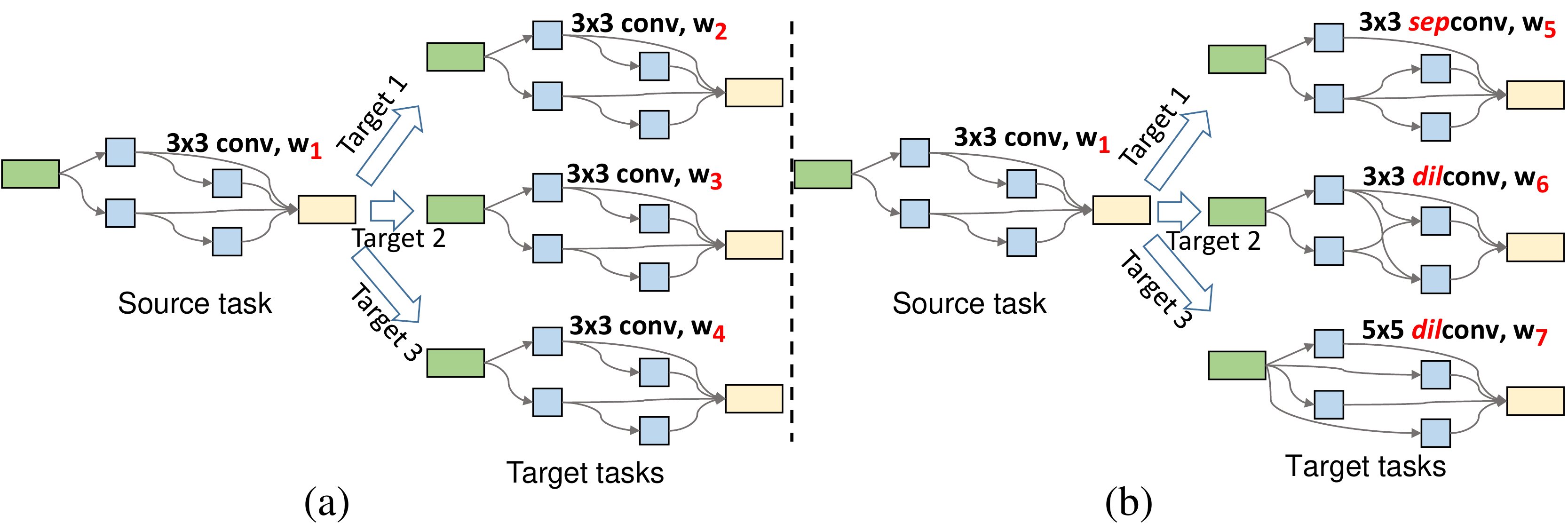}
\vspace{-5pt}
\caption{{Comparison between (a) weight pretraining and finetuning (WP\&F) and (b) the proposed framework of neural transfer and architecture adaptation (NTAA). In WP\&F, only network weights are transferred to the downstream tasks, e.g., from $w_{\textcolor{red}{1}}$ in a source task to $w_{\textcolor{red}{2}},w_{\textcolor{red}{3}},w_{\textcolor{red}{4}}$ in the target tasks. While in our NTAA, both the network weights and architecture are transferred to the downstream tasks, e.g., from $<$$\text{3x3 conv,} w_{\textcolor{red}{1}}$$>$ in a source task to $<$$\text{3x3 sepconv,} w_{\textcolor{red}{2}}$$>$, $<$$\text{3x3 dilconv,}w_{\textcolor{red}{3}}$$>$, $<$$\text{5x5 dilconv,} w_{\textcolor{red}{4}}$$>$ in the target tasks.}}\label{fig:intro}
\vspace{-11pt}
\end{figure*}

The success of ImageNet has enabled a standard paradigm of image recognition. Specifically, neural networks are often first pretrained on ImageNet to obtain a set of pretrained weights (e.g., $w_{\textcolor{red}{1}}$ in Fig. \ref{fig:intro} (a)). Then, these pretrained network weights are further finetuned on a smaller, task-specific dataset to obtained the final optimal weights (e.g., $w_{\textcolor{red}{2}},w_{\textcolor{red}{3}},w_{\textcolor{red}{4}}$ in Fig. \ref{fig:intro} (a)). Such a paradigm has led to state-of-the-art performance in almost all computer vision tasks, including person re-identification (re-ID) \cite{zhong2017random}, human attribute recognition (e.g., age estimation and gender recognition) \cite{DBLP:conf/iccv/LiuLWT15}, and image classification \cite{DBLP:conf/icml/TanL19}.

However, this paradigm of weight pretraining and finetuning is not always effective, especially when the gap between the source and target domain tasks is large. The reason behind that is three-fold. \textbf{First}, different domains with distinct recognition targets may need different levels/paths of feature hierarchy and different network topological connectivity. For example, as is known to all, DenseNet \cite{Huang2017Densely_cvpr} achieves good performance on CIFAR-10 while having poor performance on ImageNet. In contrast, ResNets \cite{he2016deep} obtain high top-1 accuracy on ImageNet but have high error on CIFAR-10. \textbf{Second}, transferred to the smaller dataset or simpler task, the architectures may be towards shallower; on the contrary, transferred to challenging tasks, the transferred architecture may be deeper. \textbf{Third}, different adaptation tasks prefer different new operations. For instance, the CIFAR-10 task likes operations that can perform feature augmentation while the person re-identification task prefers operations that capture global dependencies. In summary, the weight adaptation with the fixed network architecture suffers from the limited transfer capability from one domain into another. Usually, adapting specific neural network structures for different tasks is necessary to achieve state-of-the-arts, as Fig. 1 (b) illustrates.

{To guarantee each task a personalized network structure, recently, many efforts have been made to manually \cite{Tan2020EfficientDet_cvpr,Zhang2020Relation_cvpr} or automatically \cite{Chen2019DetNAS_arxiv,Quan2019Auto_iccv,Liu2019Auto_cvpr} design neural networks for different tasks. Among these works, neural architecture search (NAS) is well-known for searching for neural structures in an automated way \cite{nas}. However, the scheme of using NAS is quite inefficient and time-consuming \cite{Chen2019DetNAS_arxiv,Quan2019Auto_iccv,wang2021heterogeneous}, which contains three indispensable stages. The three phases include:\begin{itemize}
\item Searching for a different architecture for each target task.
\item Pretraining these architectures on ImageNet one by one.
\item Finetuning them on the target tasks one by one.
\end{itemize}These redundant phrases are caused by the isolated optimization of the network architecture and network weights in NAS.}

In this work, we propose a new transfer learning\footnote{Generally, transfer learning refers to reusing a model developed for a task as the starting point for a model on a second task. In this work, transfer learning specially represents using pre-trained neural networks on a large image classification dataset for benefiting different image recognition tasks.} framework called neural transfer and architecture adaptation (NTAA), to address the above problems. We find that dynamically tuning the network architectures tailored for each domain task along with fine-tuning weight leads to more efficient and effective transfer learning, compared to the existing adaptation pipelines that only tune the weights regardless of the architecture backbone. Given a network architecture $\alpha_0$ whose network weights have been pretrained on ImageNet, our NTAA comprises of three main steps. \textbf{i)} We design a search space of network architectures $\mathcal{A}$ such that the given architecture $\alpha_0$ can be seen as an instance in the search space $\mathcal{A}$ (i.e., $\alpha_0\in\mathcal{A}$). \textbf{ii)} We start with $\alpha_0$, and jointly finetune the network weights and network architecture within the space $\mathcal{A}$. \textbf{iii)} The target architecture $\alpha^*$ is obtained after the optimization, based on which we further finetune the network weights for some epochs. {Our method can be easily generalized to an unsupervised paradigm by training the supernet in the source domain task in a self-supervised manner and performing the linear evaluation in the downstream tasks to search for the best architecture. Since the supernet needs no fine-tuning during searching, and since the self-supervised learning features have good generalization ability \cite{He2020Momentum_cvpr}, the unsupervised version's search efficiency can be significantly higher than the supervised one.}

%\textcolor{red}{Our method can be easily generalized to an unsupervised paradigm by replacing supernet training with self-supervised learning in the source domain task. Then, the supernet can be considered as a super pretrained model, which is a substitute for the original $\alpha_0$, called $Super~\alpha_0$. $Super~\alpha_0$ is saved and is ready for all the downstream tasks without re-training in the future. Next, we perform the linear evaluation in the downstream tasks and search for the best architecture $\alpha^*$. At last, we further finetune the network weights for some epochs. Since $Super~\alpha_0$ needs no re-training and since the self-supervised learning features have good generalization ability \cite{He2020Momentum_cvpr}, the search efficiency of our method can be significantly improved.}

We further discuss the reason behind the effectiveness of our NTAA and compare it with existing NAS approaches \cite{nas,real2018regularized}. {In addition to the aforementioned search inefficiency \cite{Chen2019DetNAS_arxiv,Quan2019Auto_iccv,Liu2019Auto_cvpr}, current NAS methods are also challenged with search ineffectiveness \cite{Li2020Block_cvpr,sciuto2019evaluating,Yang2020NAS_iclr}. In fact,} NAS is far away from being broadly practical in general image recognition. Some recently proposed works \cite{sciuto2019evaluating,Yang2020NAS_iclr} suggest that the performances of some NAS methods in certain scenarios are even worse than random architecture selection. In this work, we reveal the limitation of applying current NAS solutions within the transfer learning from both the principled and empirical perspectives. Solving a NAS model, essentially, is to find a network architecture with the highest train-from-scratch accuracy. The problem is very hard due to the intolerable cost of training each architecture in the search space from scratch. To simplify the problem, existing NAS solutions often use under-trained or shared network weights to evaluate an architecture's performance by implicitly assuming the network architecture is independent of the network weights. The inappropriate evaluation has led to the unreproducible results of some existing NAS methods. On the contrary, our NTAA formulates the optimization of the network weights and architecture as a joint entity and solves the two parts synchronously. In addition to providing support to NTAA, the conducted analysis also benefits the future research of NAS.

Overall, the paper makes the following three contributions.\begin{itemize}
\item We propose a general framework to adjust the network weights along with the network architecture adaptation synchronously. In a defined search space of network architecture, our method searching for an optimal architecture reduces to selecting an appropriate operation from the candidate operation set for each layer.

\item Our method can be easily generalized to an unsupervised paradigm, which further improves the search efficiency of our method.

\item We provide principled analysis to explain why our framework works by investigating the ineffectiveness of existing solutions to NAS. We find that preserving the joint distribution of the network architecture and weights is of importance. This analysis not only benefits transfer learning but also provides insights for NAS.

\item We conduct extensive experiments on a variety of tasks, \emph{i.e.}, person re-ID, age estimation, gender recognition, image classification, and unsupervised domain adaptation. On these tasks, we achieve state-of-the-art performance, demonstrating the superiority of our learning framework.
\end{itemize}

The remainder of this work is organized as follows. We review the previous works relevant to our method in Section \ref{sect:related_work} and introduce the NTAA learning framework and its extension in Section \ref{sect:method}. In Section \ref{sect:theory}, We provide the analysis to explain the reasons of the effectiveness of NTAA in a principled way. Section \ref{sect:exp} presents the experimental results and comparisons. Section \ref{sect:conclusion} concludes this paper.

\section{Related Work}\label{sect:related_work}

\textbf{Weight Pretraining and Finetuning (WP\&F).} There is an overwhelming amount of deep-learning-based methods borrowing the powers from pre-training neural networks on large-scale datasets. Decades ago, Hinton et al., introduced transfer learning into training neural networks, especially under unsupervised learning scenarios \cite{hinton1986learning}. Transfer learning techniques attracted much interest since 2012 when large-scale datasets such as ImageNet \cite{Krizhevsky2012ImageNet_nips} were utilized in many image recognition tasks. Pre-training models on ImageNet is a key to achieve state-of-the-art performances in various tasks such as object detection \cite{ren2015faster}, semantic segmentation \cite{chen2018deeplab}, video recognition \cite{wu2018group}, person re-identification \cite{zhong2017random}, human attribute recognition \cite{DBLP:conf/iccv/LiuLWT15}, and image classification \cite{DBLP:conf/icml/TanL19,Shu2018Image_csvt,Tang2016Generalized_mcca,Shu2015Weakly_acmmm}. Moreover, transferring from ImageNet not only benefits the accuracies of the target tasks but also speeds up the learning process \cite{chen2018deeplab,Shu2018Image_csvt,Tang2016Generalized_mcca,Shu2015Weakly_acmmm}. In addition to computer vision, weight pretraining and finetuning is also used in other domains, such as natural language processing (NLP). Language model pretraining has shown remarkable improvement for many NLP tasks, such as natural language inference, paraphrasing, named entity recognition, and question answering. Notable works include ELMo \cite{Peters2018Deep_NAACL}, OpenAI GPT \cite{Radford2018improving_arxiv}, and BERT \cite{Devlin2019pretraining_naacl}. ELMo \cite{Peters2018Deep_NAACL} uses the pretrained representation as an additional feature for ensembling. The way of transfer learning in OpenAI GPT \cite{Radford2018improving_arxiv} and BERT \cite{Devlin2019pretraining_naacl} are more similar to that of the standard transfer learning in computer vision, i.e., a backbone is pretrained in the source task whose network weights are further finetuned for the target task. Compared with the feature transferring in previous works, our NTAA jointly optimizes the network architecture and neural weights. Moreover, our framework can be also combined with existing transfer learning methods.

The idea of using confidence values for operation shares the merit of adaptation factors in \cite{Tran2016Adaptive_ijcnn}. There are two differences between our confidence values for operation and the adaptation factors in \cite{Tran2016Adaptive_ijcnn}. \textbf{First}, our confidence values are associated with the selection for candidate operations, which result in different network architectures, i.e., both the network architecture and the network weights can be evolved with the confidence values changing. In contrast, in \cite{Tran2016Adaptive_ijcnn}, only the model weights can be adapted with respect to the adaptation factors. Therefore, the adaptation factor used in \cite{Tran2016Adaptive_ijcnn} is only a tool in standard transfer learning, while our confidence value is a formulation for neural architecture adaptation. \textbf{Second}, a low confidence value in our method can lead to a straightforward removal of an operation of a predefined architecture; however, a low adaptation factor still assigns a small weight for a pretrained model weight.

\textbf{Neural Architecture Search (NAS).} Recently, there has been a growing interest in developing algorithmic solutions to automate the manual process of designing a machine learning algorithm. In particular, neural architecture search (NAS) is expected to reduced the effort of human experts in network architecture design \cite{nas,liu2017progressive,xie2018snas,wu2018fbnet,cai2018proxylessnas,DBLP:journals/corr/abs-1811-09828,DBLP:journals/corr/abs-1807-11626,enas,DBLP:journals/corr/abs-1902-09635,cai2018proxylessnas,DBLP:journals/corr/abs-1904-00420}. However, there is still an unsolved problem in NAS, i.e., how to efficiently solve a NAS model. The most mathematically accurate solution is to train each of the candidate architectures within the search space from scratch and compare their performance. The architecture with the most satisfactory performance is regarded as the target architecture. However, this solution is so extremely time-consuming as the search space is usually quite large (e.g., $>1e^{20}$ \cite{Jiang2020SP_cvpr}). To cope with this problem, many works have explored to train only the candidate architectures in a search \emph{sub-space} by using reinforcement learning (RL) \cite{nas,rl17iclr} or evolution learning \cite{real2018regularized} to guide the search direction. For example, in the RL-based NAS, only the most potential candidate architectures with the largest rewards are trained because they are assumed to contain the target architecture. These NAS algorithms have achieved remarkable performance. However, they are still computationally demanding. For example, obtaining a state-of-the-art architecture for CIFAR-10 required 1,800 GPU days of reinforcement learning \cite{nas} and 3,150 GPU days of evolution \cite{real2018regularized}. This indicates that training the candidate architectures in the search \emph{sub-space} (e.g., 1 million architectures) is still impractical, as training even \emph{one} architecture costs a long time (e.g., more than 10 GPU days for a ResNet \cite{he2016deep} on ImageNet).

To speed up NAS, the current methods have given up the mathematically accurate solution. They propose not to train each of the candidate architectures from scratch. Instead, they propose to train different candidate architectures by sharing the network weights \cite{han2019once}. Notable works include the weight-sharing reinforcement learning methods \cite{enas}, attention-based differentiable methods (e.g., DARTS \cite{liu2018darts} and ProxylessNAS \cite{cai2018proxylessnas}), and supernet-based methods \cite{DBLP:journals/corr/abs-1904-00420,DBLP:journals/corr/abs-1908-06022} in which a supernet is built to represent the full search space and each path is a stand-alone model. However, there is no theoretical guarantee that weight-sharing methods should work. Actually, as is suggested by \cite{sciuto2019evaluating,Yang2020NAS_iclr}, many existing solutions to NAS are not better than random architecture selection. In our work, we explain the ineffectiveness of existing solutions to NAS in a principled way.

\textbf{Self-Supervised Learning.} Recently, unsupervised learning has recently shown remarkable progress in representation learning, especially in natural language understanding and computer vision. Notable works in natural language understanding include GPT \cite{Radford2018improving_arxiv} and BERT \cite{Devlin2019pretraining_naacl}. Among unsupervised learning, the results of self-supervised learning are most promising in the computer vision task. Specifically, self-supervised learning in computer vision can be divided into three groups, including low-level methods, mid-level methods, and high-level methods. Low-level methods include de-noising auto-encoders \cite{Vincent2008Extracting_icml}, context auto-encoders \cite{Pathak2016Context_cvpr}, or colorization \cite{Zhang2016Colorful_eccv}. Mid-level methods include patch orderings \cite{Doersch2015Unsupervised_iccv,Noroozi2016Unsupervised_eccv}. However, both low-level methods and mid-level methods have poor performance in learning universal features. The most successful methods are high-level methods, i.e., contrastive learning methods. Notable works include MoCo \cite{He2020Momentum_cvpr,Chen2020Improved_arxiv}, simCLR \cite{Chen2020Simple_arxiv}, and BYOL \cite{Grill2020Bootstrap_arxiv}. For example, on ImageNet, the top-1 accuracy of BYOL is 74.3\%, which is close to that of supervised learning, i.e., 76.4\%.

Despite the promising accuracy and high expectation, the learning efficiency of self-supervised learning is low. {Self-supervised learning usually costs ten times longer time for optimization than supervised learning.} Specifically, for the task of training a ResNet50 on ImageNet, the supervised method costs about 100 epochs, while simCLR and BYOL cost 1,000 epochs, and CoCo v2 costs 800 epochs. The inefficiency of existing contrastive learning is the unreliability of the momentum encode. Specifically, the momentum encoder in existing methods is an Exponential Moving Average (EMA) networks of the encoder networks. However, the EMA networks are not reliable in the earlier stage. Therefore, the learning efficiency of existing self-supervised learning is low. On the contrary, in our method, there is a reliable pretrained model. Using the pretrained model to replace the EMA networks in self-supervised learning significantly improves the learning efficiency.

\section{Methodology} \label{sect:method}
{We start by briefly reviewing the technical background related to our work and then introduce our learning framework in detail. An extension of our framework to an unsupervised paradigm is further discussed, followed by more implementation details. }

\subsection{Preliminaries}\label{sect:standard}

\textbf{Deep Neural Networks} A deep model is written as:\begin{small}\begin{equation}\label{eq:deep}
\begin{aligned}
&\Phi(X,W^{(1)},\cdots, W^{(i)},\cdots, W^{(K)}) \\&= \psi_K(\cdots \psi_{i}(\cdots \psi_1(X\otimes W^{(1)})\cdots \otimes W^{(i)})\cdots \otimes W^{(K)}),
\end{aligned}
\end{equation}
\end{small}where the whole neural network is formulated as a complicatedly nonlinear function $\Phi(\cdot)$. $X$ is the input. \begin{small}$W = \{W^{(1)},\cdots, W^{(i)},\cdots, W^{(K)}\}$\end{small} denotes the network weights. $K$ is the depth of the network. $\otimes$ denotes a convolution operation (e.g., vanilla convolution, separate convolution, dilated convolution). $\psi(\cdot)$ denotes a nonlinear activation function (e.g., batch normalization + ReLU). For presentation simplification, Eqn. \eqref{eq:deep} is re-written as:\begin{small}\begin{equation}\label{eq:dl}
\Phi_{W,\alpha}(X) = (C_{W^{(K)}}\cdots\circ C_{W^{(i)}}\cdots \circ C_{W^{(1)}})(X),
\end{equation}
\end{small}where $C_{W^{(i)}}$ denotes a convolution operation using the network weight $W_{(i)}$ as convolutional filters and $\circ$ denotes a sequence of operations (e.g., convolution) in a network architecture $\alpha$. In the following, we use $C_{i}$ to denote $C_{W^{(i)}}$ for notation simplification. Here, $i$ denotes the $i$-th layer. In other words, the network architecture $\alpha$ can be represented as:\begin{small}\begin{equation}\label{eq:arch}
\alpha=C_{K} \cdots \circ C_{i}\cdots \circ C_{1}.
\end{equation}
\end{small}Finally, a deep learning problem is an optimization problem in the form of:\begin{small}\begin{equation}\label{eq:loss}
\min_w \mathcal{L}(Y, \Phi_{W,\alpha}(X)) + \lambda ||W||_2
\end{equation}
\end{small}where $\mathcal{L}$ denotes the loss function. $\lambda ||W||_2$ is a weight decay term that regularizes the model to prevent overfitting. $\lambda$ is a hyper-parameter which is usually set as $1e^{-4}$.

\begin{figure*}[t]
\centering
\includegraphics[width=1.0\textwidth]{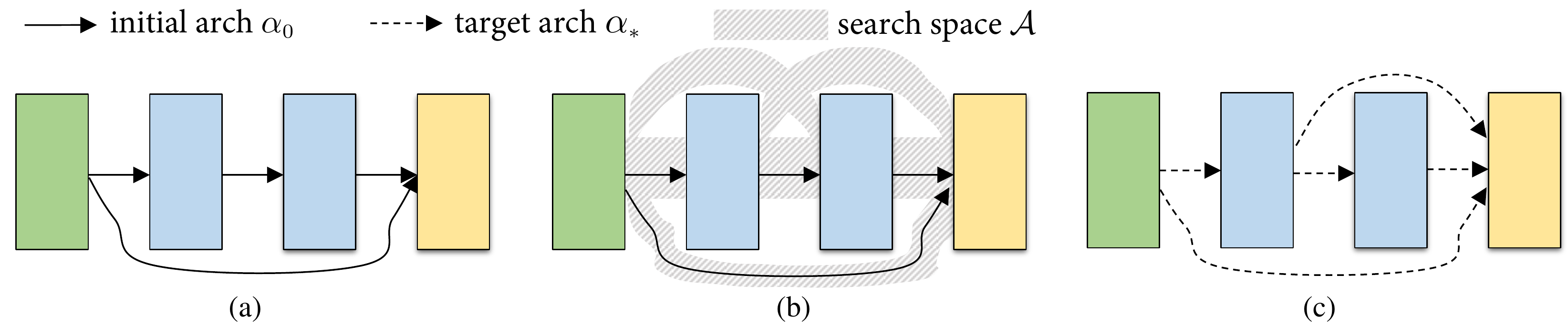}
\caption{An illustration of our framework: (a) the initial architecture; (b) the search space; (c) the target architecture.}\label{fig:method}
\vspace{-11pt}
\end{figure*}

\vspace{6pt}
\textbf{Weight Pretraining and Finetuning} Given a network $\alpha_0$, pretraining is to train $\alpha_0$ from scratch on a source domain task. After the pretraining, we obtained the pretrained network weights $W_0$. Then, a pretrained deep model can be written as \begin{small}$\alpha_0 = (C_{W_0^{(K)}}^{0}\cdots \circ C_{W_0^{(i)}}^{0} \cdots \circ C_{W_0^{(1)}}^{0})$\end{small}. Remarkably, in computer vision, ImageNet has been proved to be transferable to many other tasks. Therefore, ImageNet is considered as the most commonly used source domain task. Moreover, the pretrained ImageNet models are available online in the model zoos of almost all the deep learning frameworks (e.g., Caffe, Tensorflow, and Pytorch). Hence, the pretraining stage can be omitted in both standard transfer learning and our method.

In the paradigm of weight pretraining and finetuning, the network weights are tuned while the network architecture is fixed when transferred to a target task. The goal of weight pretraining and finetuning is to find the optimal weights $W^* = W_0 + \Delta W^*$ for a fixed architecture $\alpha_0$ which is subject to the condition that $W$ is initialized as $W_0$, i.e.,\begin{small}\begin{equation}\label{eq:standard}
\begin{aligned}
\Delta W^* &=\mathop{\arg\min}_{\forall \Delta W } \mathcal{L}(Y, \Phi_{W_0 + \Delta W,\alpha_0}(X)) + \lambda ||W_0 + \Delta W||_2\\
s.t. ~~~~ &\left\{
\begin{aligned}
\text{$W_0$ is pretrained in the source task}\\
\text{$\alpha_0$ is kept fixed}\\
\end{aligned}
\right.
\end{aligned}
\end{equation}\end{small}Note that $\alpha_o$ keep unchanged before and after the optimization, indicating that the network architecture is unlearnable during the process of weight pretraining and finetuning. 

\vspace{6pt}
\textbf{Neural Architecture Search} Neural architecture search (NAS) is an AutoML technique, which aims at automatically designing neural architectures for different tasks. NAS is not defined for a transfer learning problem, i.e., NAS does not exploit the pretrained weights obtained on a source task. Formally, NAS is formulated as:\begin{small}\begin{equation}\label{eq:nas}
\begin{aligned}
\alpha^* =\mathop{\arg\min}_{\forall \alpha } ( \mathop{\min}\limits_{\forall W}(\mathcal{L}(Y, \Phi_{W,\alpha}(X)) + \lambda ||W||_2))\\\end{aligned}
\end{equation}\end{small}As shown, for each architecture $\alpha$, we should train its network weights from scratch to convergence.

\subsection{Neural Transfer and Architecture Adaptation}\label{sect:nat}

Different from weight pretraining and finetuning, our method is supposed to synchronously adjust the network weights as well as network architecture. The formulation of our method is as follows:\begin{small}\begin{equation}\label{eq:nwat}
\begin{aligned}
\Delta W^*, \Delta\alpha^* &=\mathop{\arg\min}_{\forall \Delta W, \forall \Delta\alpha } \mathcal{L}(Y, \Phi_{W_0 + \Delta W,\alpha_0 + \Delta\alpha}(X)) \\
&+ \lambda ||W_0 + \Delta W||_2\\
s.t. ~~~~ &\left\{
\begin{aligned}
\text{$W_0$ is pretrained in the source task}\\
\text{$\alpha_0$ is predefined for the source task}\\
\end{aligned}
\right.
\end{aligned}
\end{equation}\end{small}Here, we use $\Delta\alpha$ to denote the modification between the searched architecture and the architecture predefined for the source task. We use $\alpha_0 + \Delta\alpha$ to denote making a modification $\Delta\alpha$ for the architecture $\alpha_0$. Our goal is to find optimal weights $W^* = W_0 + \Delta W^*$ and an optimal architecture $\alpha^* = \alpha_0 + \Delta\alpha^*$ which is subject to the conditions that $W$ is initialized as $W_0$ and $\alpha$ is initialized as $\alpha_0$.

As tuning the network weight is easy to understand, in the following, we only focus on the adaptation of the network architecture for notation simplification. Eqn. \eqref{eq:nwat} can be simplified as follows:\begin{small}\begin{equation}\label{eq:simple}
\begin{aligned}
\Delta\alpha^* &=\mathop{\arg\min}_{\forall \Delta\alpha } \mathcal{L}(Y, \Phi_{\alpha_0 + \Delta\alpha}(X)) \\
s.t. ~~~~ &\text{$\alpha_0$ is predefined for the source task}\\
\end{aligned}
\end{equation}\end{small}Integrating Eqn. \eqref{eq:arch} into Eqn. \eqref{eq:simple}, adapting the architecture from an initial architecture $\alpha_0$ to the target architecture $\alpha^*$ can be finalized as follows:\begin{small}\begin{equation}\label{eq:arch_adaptation}
\begin{aligned}
&\Delta C_K^*\text{,...,}\Delta C_K^*\text{,...,}\Delta C_1^* =\mathop{\arg\min}_{ \Delta C_K, \cdots,\Delta C_i\cdots, \Delta C_1 } \mathcal{L}(Y, ((C_K^0+\Delta C_K) \\
&\circ \cdots \circ (C_i^0 + \Delta C_i) \circ \cdots \circ (C_1^0 + \Delta C_1) )(X)) \\
&s.t. ~~~~ C_K^0,\cdots, C_i^0, \cdots, C_1^0 \text{are predefined for the source task}\\
\end{aligned}
\end{equation}\end{small}

\textbf{Search Space Design.} As discussed, our method is to make a set of modification $\big\{\Delta C_K, \cdots,\Delta C_i\cdots, \Delta C_1 \big\}$ for the predefined convolution operations $\big\{C_K^0,\cdots, C_i^0, \cdots, C_1^0\big\}$. Therefore, we need to design a search space for the modification $\big\{\Delta C_K, \cdots,\Delta C_i\cdots, \Delta C_1\big\}$. That is, we need to design a search space $\mathcal{A} =\{\mathcal{O}_{1}, \cdots, \mathcal{O}_{i}, \cdots, \mathcal{O}_{K}\}$ such that:\begin{small}
\begin{equation}
\Delta C_K \in \mathcal{O}_{K}, \cdots, \Delta C_i \in \mathcal{O}_{i}, \cdots, \Delta C_1 \in \mathcal{O}_{1}.
\end{equation}
\end{small}How to form the architecture search space in a reasonable way is an active problem in NAS. Conventionally, in NAS, the search space is formed by defining candidate operations for each layer. Inspired by NAS, we define the modification $C_{i}$ ($0\le i \le K$) as replacing the predefine convolution operation $C_i^0$ by another operations selected from a candidate operation set $\mathcal{O}_i$. By definition, each $\mathcal{O}^i$ consists of three types of operations:\begin{itemize}
\item{} \emph{\textbf{Convolution:} $5\times5$, $3\times3$, and $1\times1$ convolution, denoted as $o_1^i$, $o_2^i$, and $o_3^i$, respectively.}
\item{} \emph{\textbf{Pooling:} $3\times3$ max pooling, $3\times3$ average pooling, globalization operation, denoted as $o_4^i$, $o_5^i$, and $o_6^i$, respectively.}
\item{} \emph{\textbf{Others:} identity, and noise disturbing, denoted as $o_7^i$ and $o_8^i$, respectively.}
\end{itemize}where globalization indicates broadcasted global average pooling, and noise disturbing indicates adding a gaussian noise to the features. Here $i$ denotes the $i$-th layer. Note that the operations in \emph{pooling} and \emph{others} are weight-free, indicating that choosing these two types of operations makes the network shallower.

\emph{It is important to note that, different from NAS, some of the predefined convolution operations may stay unchanged after the architecture transfer. Therefore, we must ensure that $\alpha_0 \in \mathcal{A}$, i.e., all $\mathcal{O}_i$s ($0\le i \le K$) should meet the condition of $C_{i}^{0} \in \mathcal{O}_{i}$ ($0\le i \le K$)}.

Finally, searching for an optimal architecture reduces to selecting an appropriate operation from the candidate operation set for each layer. With the definition of the search space, our method is defined as an optimization problem as follows:\begin{small}\begin{equation}\label{eq:opt_arch}
\begin{aligned}
&\Delta C_K^*\text{,...,}\Delta C_K^*\text{,...,}\Delta C_1^* =\mathop{\arg\min}_{ \Delta C_K, \cdots,\Delta C_i\cdots, \Delta C_1 } \mathcal{L}(Y, ((C_K^0+\Delta C_K) \\
&\circ \cdots \circ (C_i^0 + \Delta C_i) \circ \cdots \circ (C_1^0 + \Delta C_1) )(X)) \\
&s.t. ~~~~ \left\{
\begin{aligned}
&C_K^0,\cdots, C_i^0, \cdots, C_1^0 \text{are predefined for the source task}\\
& \Delta C_i \in \mathcal{O}_{i}
\end{aligned}
\right.
\end{aligned}.
\end{equation}\end{small}Fig. \ref{fig:method} shows the relationship of the initial architecture $\alpha_0$, the target architecture $\alpha^*$, and the search space $\mathcal{A}$, in which the initial architecture $\alpha_0$ can be seen as an instance in the search space $\mathcal{A}$ (i.e., $\alpha_0 \in \mathcal{A}$).

%\textcolor{red}{Designing a larger search space can result in bad consequences in search cost and model efficiency. \textbf{First}, defining too many candidate operations can increase the search cost. For example, if we double the number of candidate operations in our search space, the required GPU memory can be beyond that our machine can be offered. \textbf{Second}, too large search space can lead to bad architecture rating, which randomizes the performance of architecture search, as is suggested by \cite{Li2020Block_cvpr}.}

{(\emph{\textbf{Rule of thumb}})} {How to design a reasonable search space is an active problem in NAS. Conventionally, in NAS, the search space is created by defining candidate operations for each layer and considering the topological connectivity of the network. These two aspects have been taken into account in our method. Among these aspects, predefining candidate operations is preferred in most of the existing NAS methods \cite{liu2017progressive,xie2018snas,wu2018fbnet,cai2018proxylessnas,DBLP:journals/corr/abs-1811-09828,DBLP:journals/corr/abs-1807-11626,enas,DBLP:journals/corr/abs-1902-09635,DBLP:journals/corr/abs-1904-00420,Jiang2020SP_cvpr,rl17iclr,han2019once,liu2018darts,DBLP:journals/corr/abs-1908-06022}. There is no instruction or guide for how to select the proper types of candidate operations. But spontaneously, current works on NAS (e.g., \cite{liu2017progressive,xie2018snas,wu2018fbnet,cai2018proxylessnas,DBLP:journals/corr/abs-1811-09828,DBLP:journals/corr/abs-1807-11626,enas,DBLP:journals/corr/abs-1902-09635,DBLP:journals/corr/abs-1904-00420,Jiang2020SP_cvpr,rl17iclr,han2019once,liu2018darts,DBLP:journals/corr/abs-1908-06022}) all consider the following three types of operations as candidates, i.e., convolution, pooling, and identity. Our search space is consistent with existing methods. Defining different search spaces can result in different possible consequences in search cost and model efficiency. \textbf{First}, defining too many candidate operations can increase the search cost. For example, if we double the number of candidate operations in our search space, the required GPU memory can be beyond what our machine can offer. \textbf{Second}, too large search space can lead to lousy architecture rating, which randomizes the performance of architecture search, as is suggested by \cite{DBLP:journals/corr/abs-1907-01845,Li2020Block_cvpr} below. \textbf{Third}, having different types of convolution can lead to different efficiency. For example, replacing vanilla convolution with MBConv can improve model efficiency but reduces the model accuracy. One possible \textbf{Rule of Rhumb} on search space design may be searching for a search space, which is an unexplored / rarely-explored area in AutoML. For example, we can first define a vast search space consisting of almost all possible spaces. Then, we divide the search space into some search sub-spaces (i.e., 1,000 subspaces). Next, we sample architectures from these sub-spaces and pre-evaluate their performance. Last, the combination of the sub-spaces achieving higher pre-evaluation accuracies is considered as our final search space.}

\vspace{6pt}
\textbf{Formulation of Architecture Optimization.} As discussed in Eqn. \eqref{eq:opt_arch}, searching for an optimal architecture reduces to selecting an appropriate operation from the candidate operation set for each layer. However, selecting an operation from the candidate set is discrete and non-differentiable, which cannot be optimized by DNNs. To address this problem, we relax the hard selection problem into a soft one. Specifically, we associate each candidate operation in $\mathcal{O}$ with a confidence value $\mathcal{P}\in [0,1]$, with $\mathcal{P}= 1$ indicating that the corresponding operation is definitely adopted. We assume that $p$ can be learned in a data-driven manner. For example, for the $i$-th layer (\begin{small}$0\le i \le K$\end{small}), the probability of selecting a 3$\times$3 convolution is defined as:\begin{small}\begin{equation}\label{eq:prob}
\begin{aligned}
&\mathcal{P}(C_i = o_2^i) = \frac{\exp(\theta_{o_2^i})}{\exp(\theta_{o_1^i})+ \exp(\theta_{o_2^i})+\cdots+\exp(\theta_{o_8^i})}\\
\end{aligned}.
\end{equation}\end{small}where $\theta_{o_2^i}$ denotes a learnable parameter measuring the probability of selecting a 3$\times$3 convolution for the $i$-th layer. The probability of selecting other operations are similarly defined. Then, the neural network as a complicated function in the searching process is defined as:\begin{equation}
 \Phi_{\theta}(X) = (\mathop{\mathbb{E}} \big[o^K\big] \circ\cdots\mathop{\mathbb{E}} \big[o^i\big]\circ \mathop{\mathbb{E}} \big[o^1\big])(X).
 \end{equation}where $\mathop{\mathbb{E}} \big[o^i\big] $ denotes an expectation of multiple choice, which is further defined as:\begin{equation}\label{eq:expectation}
 \mathop{\mathbb{E}} \big[o^i\big] = \sum\limits_{t=1}^8 \mathcal{P}(C_i = o_t^i) \big[o_t^i\big]
 \end{equation}Moreover, in Eqn. \eqref{eq:prob}, as $C_i^0$ has been pretrained to have a stronger capacity than other operation in $\mathcal{O}_i$ at the begining, we initialize the $\theta_{C_i^0}$ to be ones while initializing the other $\theta$s in $\mathcal{O}_i$ to be zeros. So far, all the learnable parameters have been introduced, including $W$ associated with the network weights and $\theta$ associated with the network architecture. Both $W$ and $\theta$ are optimized by:\begin{small}
 \begin{equation}\label{eq:measure}
 \begin{aligned}
 W^*,\theta^* = \mathop{\arg\min}_w \mathcal{L}(Y, \Phi_{W,\theta}(X)) + \lambda ||W||_2\\
 \end{aligned},
 \end{equation}
 \end{small}where $\Phi_{W,\theta}(X)$ is calculated by:\begin{equation}\label{eq:w_theta_phi}
 \Phi_{W,\theta}(X) = (\mathop{\mathbb{E}}\big[o_{W_{o^K}}^K\big] \circ\cdots\mathop{\mathbb{E}} \big[o_{W_{o^i}}^i\big]\circ \mathop{\mathbb{E}} \big[o_{W_{o^1}}^1\big])(X).
 \end{equation}where $o_{W_{o^i}}^i$ has the same meaning as $o^i$, indicating the network weights of the operation $o_{t}^i$ is denoted as $W_{o_{t}^i}$. Inspired by the weight decay term, we include a operation regularization term $(\lambda \sum_{\theta \in \alpha_0 } \theta - \lambda \sum_{\theta \in \mathcal{A}\backslash\alpha_0 } \theta)$ to regularize the network capacity. Here $\backslash$ denotes the set difference. In particular, $\lambda \sum_{\theta \in \alpha_0 } \theta$ discourages initial architectures, while $-\lambda \sum_{\theta \in \mathcal{A}\backslash\alpha_0 } \theta$ encourage the newly-introduced operations, especially the identity and the noise disturbing operations that reduce the network complexity. Finally, our learning framework is cast to an classical optimization problem:\begin{small}
\begin{equation}\label{eq:final_loss}
\begin{aligned}
 W^*,\theta^* \!\!=\!\! \mathop{\arg\min}_w \mathcal{L}(Y, \Phi_{W,\theta}(X)) + \lambda(||W||_2 + \sum_{\theta \in \alpha_0 } \theta - \sum_{\theta \in \mathcal{A}\backslash\alpha_0 } \theta),
\end{aligned}
\end{equation}
\end{small}which can be optimized using stochastic gradient descend.

\vspace{6pt}
Obtaining the target architecture $\alpha^*$ reduces to selecting an operation $C_i$ from the candidate operation set $\mathcal{O}_i$ for each layer. After searching, we have obtained $\theta$s that measures the probability of selecting an operation for the $i$-th layer. Hence, the optimal operation is the one with the maximum probability:\begin{equation}\label{eq:binary}
C_i^{*} = \mathop{\arg\max}_{\forall 1\le j \le 8} \theta_{o_j^i}.
\end{equation}The target architecture $\alpha^*$ is obtained, based on which we further finetune the network weights for some epochs.

\begin{figure*}[t]
\centering
\includegraphics[width=1.0\textwidth]{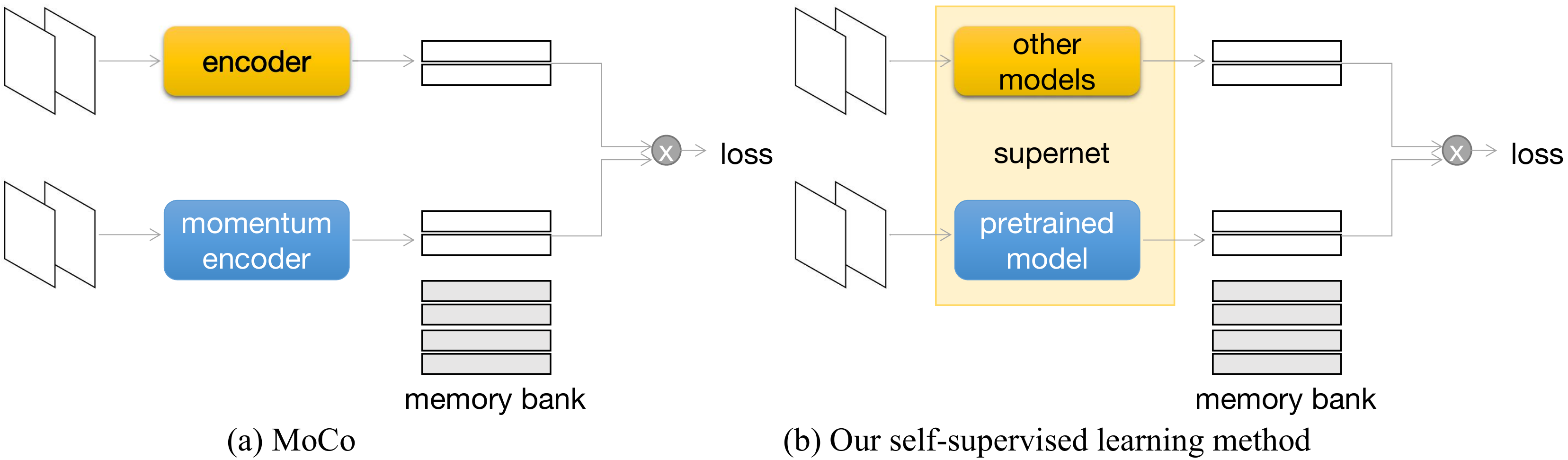}
\caption{Training $Super~\alpha$ in a self-supervised manner. (a) MoCo v1 \cite{He2020Momentum_cvpr} and v2 \cite{Chen2020Improved_arxiv}; (b) Our self-supervised learning.}\label{fig:selfsup}
\vspace{-11pt}
\end{figure*}

\subsection{Extension to Unsupervised Learning}

{Our framework can be easily generalized to a unsupervised paradigm. This needs minimal adjustment. For the presentation simplicity, we first introduce the definition of a supernet, which is a general concept widely used in the NAS community \cite{DBLP:journals/corr/abs-1904-00420,cai2018proxylessnas,liu2018darts,wu2018fbnet,xie2018snas,smash18iclr,enas}. Formally, a supernet is a directed acyclic super-graph covering a whole search space with each node representing the feature maps and each edge representing a connection between the nodes with a particular operation (e.g., a convolution). Each subnet in the supernet represents a candidate architecture in the search space}.

{With the supernet definition,  our supervised NTAA described in the previous section can be summarized into three steps. \textbf{Step 1:} A search space $\mathcal{A}$ is designed such that the given architecture $\alpha_0$ can be seen as an instance in the search space $\mathcal{A}$ (i.e., $\alpha_0\in\mathcal{A}$). This indicates we design a supernet containing $\alpha_0$ as its subnet. \textbf{Step 2:} We start with $\alpha_0$, and jointly finetune the network weights and network architecture within the space $\mathcal{A}$. This indicates we train the supernet in which the subnet $\alpha_0$ has been pretrained in the source task. \textbf{Step 3:} The target architecture $\alpha^*$ is obtained after the optimization, based on which we further finetune the network weights for some epochs. This indicates we obtain an optimal subnet from the supernet and finetune its network weights.}

{We can extend our supervised architecture search to a fast architecture search for different tasks by slightly modifying Step 2. The \textbf{NEW Step 2} is as follows: Before jointly optimizing the network architecture and network weights as supervised NTAA does, we first train the supernet in a unsupervised manner to obtain universal representations \cite{He2020Momentum_cvpr}. This universal supernet, called $Super~\alpha_0$, is used to replace the original $\alpha_0$. It is saved and is ready for all the downstream tasks without re-training in the future. Then, we perform the linear evaluation in the downstream tasks to jointly optimize the network architecture and the head weights, i.e., we initialize all $\theta$s in Eqn. \eqref{eq:prob} to be ones and freeze the backbone network weights, only learning $\theta$s and the network weights in the last layer (e.g., the ten-class classifier weights for CIFAR-10).}

{Except for Step 2, Steps 1 and 3 in the unsupervised NTAA are the same as those in the supervised NTAA without modification. For example, we perform Step 3 to search for the best architecture $\alpha^*$ and further finetune the network weights for some epochs.}

Our method of training the supernet in a self-supervised manner is novel and is different from all of the existing self-supervised learning methods. Our self-supervised learning method fits our NTAA perfectly. Specifically, the momentum encoder in existing methods, such as MoCo v1 \cite{He2020Momentum_cvpr}, MoCo v2 \cite{Chen2020Improved_arxiv}, and BYOL \cite{Grill2020Bootstrap_arxiv} are all Exponential Moving Average (EMA) networks of the encoder networks. However, the EMA networks are not reliable in the earlier stage (see Figure \ref{fig:selfsup}). Therefore, the learning efficiency of existing self-supervised learning is low. On the contrary, in transfer learning, there is a reliable pretrained model. Using the pretrained model to replace the EMA networks in self-supervised learning significantly improves the learning efficiency. Following \cite{He2020Momentum_cvpr} and \cite{Chen2020Improved_arxiv}, our loss function is also defined as InfoNCE:\begin{small}\begin{equation}
\mathcal{L}_{q, k^+, \{k^-\}} = -\log\frac{\exp(q\cdot k^+/\tau)}{\exp(q\cdot k^+/\tau)+\sum\limits_{k^-}\exp(q\cdot k^-/\tau)},
\end{equation}\end{small}where $q$ is a query feature, $k^+$ is the feature of a positive sample, and $\{k^-\}$ are the features of the negative samples.

\begin{remark}
\emph{Our unsupervised NTAA can improve search efficiency due to two reasons. \textbf{First}, Our $Super~\alpha_0$ is saved and is ready for all the downstream tasks without re-training. That is, when performing architecture search for a specific task, we can reuse the trained supernet. As most of the search time of NAS lies in the supernet training, our method significantly reduces the time's cost of architecture search for different downstream tasks. \textbf{Second}, the self-supervised learning features has good generalization ability \cite{He2020Momentum_cvpr}, which accelerates the linear evaluation and the final finetuning.}
\end{remark}

\subsection{Implementation Details}

Conventionally, in NAS, the search space is formed by defining candidate operations for each layer and considering the topological connectivity of the network. These two aspects have been taken into account in our method. In section \ref{sect:nat}, we have discussed the details of defining candidate operations for each layer. For ease of understanding, we have omitted the topological network connectivity in the search space $\mathcal{A}$. Actually, the number of layers and connection between units have been taken into account in our method.\begin{itemize}
\item There is a candidate skip connection in our search space. Specifically, as Fig. 2 (b) shows, in our search space $\mathcal{A}$, a skip layer operation is also allowed to enrich the search space. Actually, each layer is supposed to be connected with preceding three layers, i.e., both $\alpha=C_{4}\circ C_{3}\circ C_{2} \circ C_{1}$ and $\alpha=C_{4}\circ C_{1}$, $\alpha=C_{3}\circ C_{1}$ are allowed.
there is a candidate skip connection in our search space (Fig. 2 (b)). Therefore, the topological connectivity and the number of layers have also been taken into account in our architecture search.
\item Our candidate operation sets includes \emph{Pooling}, \emph{identity}, and \emph{noise disturbing}. Note that these operations are weight-free, indicating that choosing these operations can reduce the depth of the searched architecture. Therefore, the depth of our searched architecture can be further made multifarious.
\end{itemize}

\section{Principled Analysis on the Effectiveness}\label{sect:theory}

In the following, we provide the principled analysis to explain the reason behind the effectiveness of our NTAA by comparing it with the existing NAS approaches. Here, we start by discussing the defect of NAS and the ineffectiveness of directly applying NAS methods in the transfer learning.

\subsection{Ineffectiveness of NAS}

We have presented the formulation of NAS in Eqn. \eqref{eq:nas}, from which we can find that the challenge of NAS lies in the inner term of Eqn. \eqref{eq:nas}, i.e., $\mathop{\min}\limits_{\forall W} (\mathcal{L}(Y, \Phi_{W,\alpha}(X)) + \lambda ||W||_2)$, which requires us to train each of the candidate architectures in the search space $\mathcal{A}$ from scratch to convergence to obtain the optimal $W^*$. However, this is infeasible as training even one ResNet costs 10 GPU days, not to mention there are more than $1e^{20}$ architectures in the search space \cite{Jiang2020SP_cvpr}. To cope with this problem, {under-training \cite{nas,rl17iclr} and co-training \cite{Li2020Block_cvpr,DBLP:journals/corr/abs-1904-00420,cai2018proxylessnas,liu2018darts,wu2018fbnet,xie2018snas,smash18iclr,enas}} have been proposed. Under-training is to train each architecture for a few epochs (e.g., five epochs) and use the performance at the 5th epoch to evaluate the architecture. Such architecture rating is inaccurate due to the insufficient training. Therefore, co-training is proposed. Co-training is to train different candidate architectures by sharing the network weights in a supernet. Currently, co-training is the most commonly-used method in NAS. However, co-training is still ineffective because the architecture rating based on the subnet disengaged from supernet {is inaccurate \cite{Li2020Block_cvpr,sciuto2019evaluating,Yang2020NAS_iclr,anonymous2021exploring}}.

In the following, we analyze the inaccurate rating of weight-sharing NAS by using the tool of generalization boundedness.
\begin{theorem}\label{theorem:bound}
\textbf{(Generalization boundedness).} Let $\eta$ be the learning rate, $\mathcal{S}$ be the number of the candidate operations for each layer, and $T$ be the training iterations. Let $\mathcal{R}$ be the misclassification error, $L$ be the loss function, and $\mathbf{W}$ be the learnable parameters. Then, for any candidate architecture $j$, the misclassification error is upper bounded by:\begin{equation}\label{eqn:bound}
\min\limits_{t\in\big[T\big] } \mathbb{E} \big[\mathcal{R}(\mathbf{W}_{t,j})\big] \le  \mathcal{O}\Big(\frac{\ln^{2}(T)+\ln(\mathcal{S}T)}{T}\Big).
\end{equation}
\end{theorem}

\emph{Proof sketch.} Since the expected loss is independent from sampled data, we have: \begin{equation}\label{eqn:time}
\mathop{\mathbb{E}}\limits_{\mathcal{D}_{t+1}} \big[\mathcal{L} (\mathbf{W}_t)\big] = \mathop{\mathbb{E}}\limits_{\mathcal{D}_{t}, \mathcal{B}_t} \big[\mathcal{L}(\mathbf{W}_t)\big] = \mathop{\mathbb{E}}\limits_{\mathcal{D}_{t}} \big[\mathcal{L}(\mathbf{W}_t)\big].
\end{equation}As the loss upper-bounds the error rate \cite{MianjyConvergence_neurips}, we have:\begin{equation}
\begin{aligned}
&\mathop{\mathbb{E}}\limits_{\mathbf{W}_{1}, \mathcal{D}_T} \big[ \frac{1}{T} \sum\limits_{t=1}^{T} \mathcal{R} (\mathbf{W}_t) \big] \le \mathop{\mathbb{E}}\limits_{\mathbf{W}_t, \mathcal{D}_T} \big[\frac{1}{T} \sum\limits_{t=1}^{T} \mathcal{L} (\mathbf{W}_t) \big]\\
&\le \mathop{\mathbb{E}}\limits_{\mathbf{W}_t, \mathcal{D}_T, \mathcal{B}_T} \big[\frac{1}{T} \sum\limits_{t=1}^{T} \mathcal{L}_t(\mathbf{W}_t) \big]  \text{~~~~(\emph{by Eqn. \eqref{eqn:time}})}\\
& \le \frac{\mathbb{E}_{\mathbf{W}_1} \big[ \| \mathbf{W}_1 - U\|_F^2 \big] }{\eta T} + \frac{2}{T}\sum\limits_{t=1}^{T} \mathop{\mathbb{E}}\limits_{\mathbf{W}_1, \mathcal{D}_T, \mathcal{B}_T}\big[ \mathcal{L}_{t}^{(t)} (U)\big].
\end{aligned}
\end{equation}Let $U= \mathbf{W}_1 -\lambda V$ with $\|V\|_F \le 1$ and $\lambda=c_1 \ln (2 \eta T) + \sqrt{c_2\ln (c_3\eta\sqrt{\mathcal{S}} T^2) }$ with $c_1$, $c_2$, and $c_3$ being constants. Since $\|\mathbf{W}_1 - U\|_F^2 = \|\mathbf{W}_1 - \mathbf{W}_1 - \lambda V\|_F^2 = \lambda^2 \| V\|_F^2 \le \lambda^2$, we have: $\frac{\mathbb{E}_{\mathbf{W}_1} \big[ \| \mathbf{W}_1 - U\|_F^2 \big] }{\eta T} \le \frac{\lambda^2}{\eta T}$. Since $\mathcal{L}_t^{(t)} (U) \le \frac{\lambda^2}{2 \eta T}$ and $\mathcal{L}_t^{(t)} (U) \le c_4\sqrt{\mathcal{S}}$, we have: $\mathbb{E} \big[ \mathcal{L}_t^{(t)} (U) \big] \le (1 - \sigma) \frac{\lambda^2}{2 \eta T}  + \sigma c_4\sqrt{\mathcal{S}} \le \frac{\lambda^2}{2\eta T} + \sigma c_4\sqrt{\mathcal{S}} \le \frac{\lambda^2}{2\eta T} + \frac{1}{2\eta T} \le \frac{\lambda^2}{\eta T}$, where $\sigma \triangleq \frac{1}{2 c_4\sqrt{\mathcal{S}}\eta T}$. This finally yields:\begin{equation}
\min\limits_{t\in\big[T\big] } \mathcal{R}(\mathbf{W}_{t,j})\le \frac{3\lambda^2}{\eta T} \le \mathcal{O}\Big(\frac{\ln^{2}(T)+\ln(\mathcal{S}T)}{T}\Big),
\end{equation}which completes the proof.

%\begin{itemize}
%\item By Lemma 5, with probability at least $1-\sigma$, it holds that $\mathcal{L}_t^{(t)} (U) \le \frac{\lambda^2}{2 \eta T} \triangleq u_1$.
%\item By Lemma B, it holds with probability one that $\mathcal{L}_t^{(t)} (U) \le \frac{c\sqrt{m}}{\ln (2)} +1 \le 2c\sqrt{m} \triangleq u_2 $.
%\end{itemize}

\begin{remark}
\textbf{(Inaccurate rating in existing NAS).} \emph{Theorem \ref{theorem:bound} shows that the misclassification error (usually measuring the generalization ability) of the weight-sharing solution has an upper bound related to the search space size (i.e., $\mathcal{S}$). This might provide an explanation for the inaccurate rating problem in existing NAS. Specifically, for the models learned by the weight-sharing solution, increasing the search space size (i.e., $\mathcal{S}$) would lead to a larger upper bound of the misclassification error and thus the poorer generalization ability.}

\vspace{3pt}

\emph{A poor generalization ability implies that the architecture rating based on the weight-sharing solution might not be predictive of the true rating of a neural architecture. Suppose an architecture has a good ground-truth rating, but its generalization ability based on the weight-sharing solution is poor, and thus its validation accuracy is low. Then, this architecture might have a bad predicted rating.}

\vspace{3pt}

\emph{In summary, the large search space might be the cause of the ineffectiveness of existing NAS because a large search space would result in a poor generalization ability of the weight-sharing solution and further would lead to the inaccurate rating of the network architecture, which finally leads to the ineffectiveness of NAS.}
\end{remark}

\subsection{Effectiveness of NTAA}

By comparing Eqn. \eqref{eq:nas} and Eqn. \eqref{eq:nwat}, we reveal that a NAS model aims at searching for an optimal architecture that can be trained from scratch. Therefore, good architecture rating is necessary to ensure the ability of a network to be trained from scratch. Differently, our NTAA synchronously optimizes the network weights and network architecture, which needs no architecture rating.

Generally, NTAA is better compatible with the transfer learning scenarios in a principled way compared with traditional NAS solutions. In particular, Eqn. \eqref{eq:nwat} shows that our NTAA combines the network weights and architecture learning with a joint distribution, and optimizes them synchronously. After the joint optimization, an optimal pair \{$architecture$ $(\alpha_{*})$, $weights$ $(W_{*})$\} indicating a good adaptation for transfer learning tasks is obtained. Note that obtaining \{$\alpha_{*}$, $W_{*}$\} in Eqn. \eqref{eq:nwat} does not necessarily indicate obtaining an optimal solution to NAS in Eqn. \eqref{eq:nas}. Specifically, although \{$\alpha_{*}$, $W_{*}$\} has good accuracy in transfer learning using our NTAA, it may not be obtained by NAS method, because $W_{*}$ may be inaccessible when the architecture $\alpha_{*}$ is trained from scratch. Actually, training from scratch is required by the definition of NAS as Eqn. \eqref{eq:nas} suggests. The inaccessibility of $W_{*}$ has also been discussed in the theory of lottery ticket hypothesis \cite{Frankle2019lottery_ICLR,SoelenS2019using_IJCNN,Morcos2019One_CoRR}.

\subsection{Advantage of NTAA from the Practical Perspective}

The above analysis has shown the advantage of our NTAA over existing NAS methods. From the practical perspective, our NTAA aims at improving the fundamental and universal problem of standard transfer learning, while NAS aims at searching for an effective architecture for different tasks. Limited by the number of training examples and the huge cost of annotation, many tasks (e.g., person re-ID, age estimation, gender recognition, and image classification) should be first trained on large-scale datasets (e.g., ImageNet) to obtain transferable representations and subsequently fine-tuned on the specific task. Directly searching architectures for these tasks lead to a suboptimal solution. On the contrary, our NTAA exploits both the power of standard transfer learning and NAS, achieving a better performance.

\section{Experiments}\label{sect:exp}

To justify the effectiveness of NTAA, we have conducted extensive experiments on a wide range of computer vision tasks such as person re-ID, age estimation, gender recognition, image classification, and domain adaptation. Our goal is to tune the standard backbone, including ResNet50 \cite{he2016deep}, ResNet101 \cite{he2016deep}, InceptionV2 \cite{szegedy2015going}, VGG \cite{simonyan2014very}, AlexNet \cite{Krizhevsky2012ImageNet_nips}, and the recently proposed EfficientNet \cite{DBLP:conf/icml/TanL19}.

\subsection{Comparison with Weight Pretraining and Finetuning in Person Re-Identification}

We first evaluate the effectiveness of our NTAA in the well-known transfer learning task, i.e., re-ID, which has been extensively studied in recent years \cite{farenzena2010person,gray2008viewpoint}. It refers to the problem of re-identifying individuals across cameras. Solving re-ID problems is very challenging but has many applications in video surveillance for public safety. Recent state-of-the-art re-ID models are all built upon standard transfer learning techniques \cite{li2017learning,zhao2017deeply,bai2017scalable,sun2017svdnet,zheng2017unlabeled,su2017pose,hermans2017defense,zhong2017random}. Specifically, a standard ResNet-50 is first pretrained on ImageNet, and then the network weights are tuned to the re-ID domain for adaptation. For the evaluation, the test set is further divided into a gallery set of images and a probe set. We use the Rank-1, Rank-5, Rank-10, and mAP as the evaluation metric, which are standard metrics in re-ID. Note that all the compared methods only reported the Rank-1 and mAP metric on CUHK03. We are not able to access to their Rank-5 and Rank-10 results. Therefore, we only compare the Rank-1 and mAP results on CUHK03.

\begin{table}[t]
\small
\caption{Comparison on a Person Re-identification task (Market-1501). (\textbf{WP\&F:} weight pretraining and finetuning; \textbf{SOTA:} state of the art; \textbf{R-1,5,10:} Rank-1,5,10; \textbf{+E:} random erasing; \textbf{+R}: re-ranking; \textbf{UNTAA:} unsupervised NTAA)}\label{tab:market-1501}
\centering
\begin{tabular}{c|c|c|c|c|c}
Transfer       & \multirow{2}{*}{Methods}  & \multirow{2}{*}{R-1} & \multirow{2}{*}{R-5} & \multirow{2}{*}{R-10} & \multirow{2}{*}{mAP}\\ 
 type  &   &  &  &  &  \\\hline
                  & MSCAN \cite{li2017learning} & 80.31 &n/a & n/a & 57.53 \\
                  & DF \cite{zhao2017deeply} & 81.0 & n/a & n/a & 63.4 \\
                  & SSM \cite{bai2017scalable} & 82.21 & n/a & n/a & 68.80 \\
       & SVDNet \cite{sun2017svdnet} & 82.3 & n/a & n/a & 62.1 \\
    & GAN \cite{zheng2017unlabeled} & 83.97 & n/a & n/a & 66.07\\
       & PDF \cite{su2017pose} & 84.14 & n/a & n/a & 63.41 \\
                  & TriNet+E+R \cite{hermans2017defense} & 86.67 & 93.38 & n/a & 81.07 \\
                   & Omin \cite{zhou2019omni} & 94.8 & n/a & n/a & 84.9 \\
                   & JointDG \cite{zheng2019joint} & 94.8 & n/a & n/a & \textbf{{86.0}}\\
                   & IANet \cite{Hou2019Interaction_cvpr} & 94.4 & n/a & n/a & 83.1\\
\multirow{2}{*}{WP\&F} & CASN+PCB \cite{Zheng2019reid_cvpr} & 94.4 & n/a & n/a & 82.8\\
                  & CAMA \cite{Yang2019Towards_cvpr} & 94.7 & 98.1 & n/a & 84.5 \\
\multirow{1}{*}{(SOTA)} & MHN-6 \cite{Chen2019Mixed_iccv} & 95.1 & 98.1 & n/a & 85.0 \\
                   & AANet \cite{Tay2019AANet_cvpr} & 93.9 & n/a & n/a & 83.4\\
                   & $P^2$-Net \cite{Guo2019Beyond_iccv} & 95.2 & 98.2 & n/a & 85.6 \\
                   & PGFA \cite{Miao2019Pose_iccv} & 91.2 & n/a & n/a & 76.8 \\
                   & ISP \cite{Zhu2020Identity_eccv} & 95.3 & \textbf{{98.6}} & n/a & 88.6 \\
                   & CBN \cite{Zhuang2020Rethinking_eccv} & 94.3 & 97.9 & 98.7 & 83.6 \\
                   & SNR \cite{Jin2020Style_cvpr} & 94.4 & n/a & n/a & 84.7 \\
                   & $M^3$+R50 \cite{Zhou2020Online_cvpr} & 95.4 & n/a & n/a & 82.6\\                
                   & PCB \cite{sun2018beyond} & 91.2 & 96.5 & 97.3 & 76.7 \\ \hline
\multirow{3}{*}{NTAA} & PCB + NTAA & 94.5 & 97.9 & 98.3 & 84.8 \\
                 & PCB + UNTAA & 94.5 & 98.0 & 98.4 & 85.0 \\
                & NTAA + SE & \textbf{{95.6}} & 98.2 & \textbf{{98.9}} & 85.6 \\
\end{tabular}
\vspace{-11pt}
\end{table}

\textbf{Market-1501} We first conduct experiments on the Market-1501 \cite{zheng2015scalable}, which is one of the largest databases for re-ID. This database contains 32,668 images of 1,501 pedestrians captured from 6 different cameras. The dataset is split into two parts: 12,936 images with 751 identities for training and 19,732 images with 750 identities for testing. In testing, 3,368 hand-drawn images with 750 identities are used as probe set to identify the correct identities on the testing set.

\emph{Comparison with weight pretraining and finetuning.} In Table \ref{tab:market-1501}, we compare with 21 representative methods of weight pretraining and finetuning, which is also the current best models on Market-1501, including
MSCAN \cite{li2017learning}, DF \cite{zhao2017deeply}, SSM \cite{bai2017scalable}, SVDNet \cite{sun2017svdnet}, GAN \cite{zheng2017unlabeled}, PDF \cite{su2017pose}, TriNet \cite{hermans2017defense}, TriNet + Era. + Re-ranking \cite{zhong2017random}, PCB \cite{sun2018beyond}, Omin \cite{zhou2019omni}, JointDG \cite{zheng2019joint}, IANet \cite{Hou2019Interaction_cvpr}, CASN+PCB \cite{Zheng2019reid_cvpr}, CAMA \cite{Yang2019Towards_cvpr}, MHN-6 \cite{Chen2019Mixed_iccv}, AANet \cite{Tay2019AANet_cvpr}, $P^2$-Net \cite{Guo2019Beyond_iccv}, PGFA \cite{Miao2019Pose_iccv}, ISP \cite{Zhu2020Identity_eccv}, CBN \cite{Zhuang2020Rethinking_eccv}, SNR \cite{Jin2020Style_cvpr}, and $M^3$+ResNet50 \cite{Zhou2020Online_cvpr}. All settings of the above methods are consistent with the common training settings. Our NTAA achieves a rank-1 accuracy of 94.5\% with an improvement of 3.3\% over its baseline. It also surpasses other competitors by a clear margin. As Omin \cite{zhou2019omni} introduces an Omni-Scale features which contains multi-scale feature and SE global dependencies \cite{DBLP:conf/cvpr/HuSS18}, for fair comparison, we also equip our search architecture with global SE features. The results in Table \ref{tab:market-1501} show that our NTAA outperforms OSNet by a large margin. This comparison verifies the effectiveness of NTAA on re-ID tasks.

\begin{table}[t]
\small
\caption{Comparison on a Person Re-identification task (CUHK03) (\textbf{bs:} batch size; \textbf{WP\&F:} weight pretraining and finetuning; \textbf{SOTA:} state of the art; \textbf{R-1:} Rank-1; \textbf{+E:} random erasing; \textbf{+R}: re-ranking; \textbf{UNTAA:} unsupervised NTAA)}\label{tab:cuhk03}
\centering
\begin{tabular}{c|c|c|c}
Transfer Type      & Methods                              & R-1 & mAP \\ \hline
         & SVDNet \cite{sun2017svdnet}            & 41.5 & 37.3 \\
         & Omin \cite{zhou2019omni}                 & 72.3 & 67.8\\
        & TriNet + Era. \cite{zhong2017random}       & 55.5 & 50.74 \\
       & Omin \cite{zhou2019omni} & 72.3 & 67.8 \\
       & Auto-ReID \cite{Quan2019Auto_iccv}       & 73.3 & 69.3 \\
 \multirow{2}{*}{WP\&F}   & CASN+PCB \cite{Zheng2019reid_cvpr} & 71.5 & 64.4\\
                 & CAMA \cite{Yang2019Towards_cvpr} & 66.6 & 64.2 \\
\multirow{1}{*}{(SOTA)}  & MHN-6 \cite{Chen2019Mixed_iccv} & 71.7 & 65.4\\
                       & $M^3$+R50 \cite{Zhou2020Online_cvpr} & 66.9 & 60.7 \\
      & TriNet+E (Our reproduction)                  & 62.0 & 57.6 \\
                 & TriNet+E+R (bs = 32)                 & 61.2 & 55.4 \\\hline
\multirow{5}{*}{NTAA} & TriNet+E+NTAA          & 64.6 & 61.5 \\
                 & TriNet+E+UNTAA          & 66.1 & 63.2 \\
                  & TriNet+E+R+NTAA(bs = 32) & 65.0 & 61.8 \\
                  & TriNet+E+R+NTAA + longer & 71.6 & 66.1\\
                  & NTAA + SE & \textbf{{74.8}} & \textbf{{69.8}} \\
\end{tabular}
\vspace{-11pt}
\end{table}

\textbf{CUHK03} We further conduct experiments on the CUHK03 dataset \cite{li2014deepreid}, which is another of the largest databases for re-ID. This database contains 14,096 images of 1,467 pedestrians. Each person is observed by two disjoint camera views and is shown in 4.8 images on average in each view. We follow the 767/700-split setting of using CUHK03 \cite{zhong2017random}, where 767 individuals are regarded as the training set, and another 700 individuals are considered as the testing set without overlap.

\emph{Comparison with weight pretraining and finetuning.} Actually, all the state-of-the-art models in the re-ID domain uses the technique of weight pretraining and finetuning. Therefore, comparing with the methods of weight pretraining and finetuning is equivalent to compare with the current best models on CUHK03. In Table \ref{tab:cuhk03}, we compare with ten representative standard transfer learning methods, including Omni \cite{zhou2019omni}, SVDNet \cite{sun2017svdnet}, TriNet + Era. \cite{zhong2017random}, TriNet + Era. + Reranking \cite{zhong2017random}, Auto-ReID \cite{Quan2019Auto_iccv}, CASN+PCB \cite{Zheng2019reid_cvpr}, CAMA \cite{Yang2019Towards_cvpr}, MHN-6 \cite{Chen2019Mixed_iccv} , and $M^3$+ResNet50 \cite{Zhou2020Online_cvpr}. All the settings of the above methods are consistent with the common training settings. Our NTAA has achieved a new benchmarking state of the art. Specifically, NTAA achieves a rank-1 accuracy of 74.8\%. We also highlight NTAA surpasses its baseline by a clear margin (e.g., 65.0\% vs. 61.2\%). This verifies the effectiveness of NTAA on re-ID tasks.

\textbf{Results of Unsupervised Paradigm} We also report the performance of our unsupervised NTAA transfer on both Market-1501 and CUHK03. As shown in Table \ref{tab:market-1501} and Table \ref{tab:cuhk03}, our unsupervised NTAA obtains slightly higher accuracy than the supervised NTAA. For example, the rank-1 accuracies of the unsupervised NTAA and supervised NTAA on CUHK03 are 66.1\% and 64.6\%, respectively. The improvement may be attributed to the universal features learned by self-supervised learning. This comparison verifies the effectiveness of our unsupervised NTAA in addressing transfer learning problem.

\textbf{Computational Complexity} There are two rounds of running in our framework, each of which contains 100 epochs. In the first round, our goal is to tune the architecture. For the supervised paradigm, the search takes 5 hours on a sole GPU on Market-1501. In the second round, our goal is to finetune the network weights based on the optimized architecture. This round takes 3 hours on a single GPU.

For the unsupervised paradigm, the search stage takes 2 hours on a sole GPU on Market-1501, which cuts the search time of the supervised paradigm by half. This is due to two reasons. \textbf{First}, Our $Super$ pretrained model is saved and is ready for all the downstream tasks without retraining. That is, when performing architecture search for a specific task, we can reuse the trained supernet. As most of the search time lies in the supernet training, our method significantly reduces the time's cost of architecture search for the downstream tasks. \textbf{Second}, the self-supervised learning features has good generalization ability \cite{He2020Momentum_cvpr}, which accelerates the linear evaluation.

\subsection{Effectiveness Analysis in Person Re-Identification} \label{sect:empirical}

In the following, we conduct empirical studies to verify the effectiveness of our NTAA compared with existing NAS approaches.

We use two representative NAS methods as the competitors, including DARTS \cite{liu2018darts} and single-path-one-shot (SPOS) method \cite{DBLP:journals/corr/abs-1904-00420}. These two methods are the current state-of-the-art methods, which reports the best performance on ImageNet with a considerably acceptable computational cost. DARTS is an attention-based and differentiable method that densely connects the candidate operations with attention scores during the search process. Then the edges with weak attention are removed after searching, forming the target architecture. SPOS uses a supernet to represent the full search space, and each path is a stand-alone model. All the experiments in this section are conducted on Market-1501. We have the following three comparisons.

\textbf{Effectiveness / ineffectiveness of NAS.} We evaluate the effectiveness of NAS in the same search space as our NTAA mentioned in Section \ref{sect:nat}. We evaluate our method and others with the following setting. i) We randomly sample four architectures from search space with a uniform distribution and train them from scratch. The accuracies are averaged to represent the capacity of random architecture selection. ii) We use DARTS and SPOS to search for the target architectures in the search space, respectively. Then, the search architectures are trained from scratch. Note that here, our goal is to analyze the effectiveness/ineffectiveness of NAS but NOT to push the state-of-the-art. Therefore, all models have not been pretrained by ImageNet.

\begin{table}[t]
\small
\caption{Effectiveness / ineffectiveness of NAS on Market-1501. (\textbf{R-1,5,10:} Rank-1,5,10)}\label{tab:effectiveness_nas}
\centering
\begin{tabular}{c|c|c|c|c|c}
Type       & Methods                              & R-1 & R-5 & R-10 & mAP \\ \hline
\multirow{2}{*}{NAS} & DARTS & 91.7 & \textbf{{96.9}} & 97.5 & 77.2 \\
                  & SPOS & \textbf{{91.9}} & 96.9 & \textbf{{97.7}} & \textbf{{77.3}} \\ \hline
\multirow{1}{*}{Random Selection}& - & 91.6 & 96.8 & 97.5 & 77.2 \\
\end{tabular}
\vspace{-11pt}
\end{table}

The results in Table \ref{tab:effectiveness_nas} provides three insights. First, the accuracies of the NAS models are not significantly better than that of random architecture selection, indicating that the solution to NAS maybe not effective. In particular, the accuracies of DARTS and random selection are similar, while SPOS obtains slightly better results than both DARTS and random architecture selection. Considering that NAS models have spent lots of computational costs to search for the architecture, we regard the solution to NAS in this problem as ineffective. Second, although SPOS has made several efforts to improve the shared weights, the accuracy of SPOS is just slightly better than that of DARTS. This indicates that improving the solution to NAS has a long way to go. Third, the accuracies in Table \ref{tab:effectiveness_nas} are significantly lower than that in Table \ref{tab:market-1501}. The degradation is due to the lack of ImageNet pretraining, verifying the importance of exploiting both the network weights and network architecture as our NTAA does.

\begin{table}[t]
\small
\caption{NAS \emph{v.s.} NTAA for transfer learning on Market-1501. (\textbf{R-1,5,10:} Rank-1,5,10)}\label{tab:nas_nat}
\centering
\begin{tabular}{c|c|c|c|c|c}
Type       & Methods               & R-1 & R-5 & R-10 & mAP\\ \hline
\multirow{2}{*}{NAS + transfer} & DARTS & 93.2 & 97.4 & 98.2 & 81.1 \\
                  & SPOS & 93.4 & 97.6 & \textbf{{98.3}} & 81.6 \\ \hline
\multirow{1}{*}{NTAA}& - & \textbf{{94.5}} & \textbf{{97.9}} & 98.3 & \textbf{{84.8}} \\
\end{tabular}
\vspace{-11pt}
\end{table}

\textbf{Superiority of NTAA over NAS.} In addition, we compare the accuracies of NAS and NTAA on Market-1501. In our method, the architecture has first exploited the ImageNet pretraining, and then the architecture and the weights are jointly optimized in the downstream task. However, ImageNet pretraining is absent in the NAS counterparts. In the NAS counterparts, the architecture is searched in the downstream task, and then the weights are trained in the downstream task. For a fair comparison, after the architecture search in the NAS counterparts, we perform ImageNet pretraining for the searched architecture before finetuning the weights in the downstream task (i.e., Market-1501) following using the standard transfer learning pipeline \cite{li2017learning,zhao2017deeply,bai2017scalable,sun2017svdnet,zheng2017unlabeled,su2017pose,hermans2017defense,zhong2017random,sun2018beyond}. The results are reported in Table \ref{tab:nas_nat}. As shown, our NTAA obtains a 1.1\% higher accuracy than NAS models ( 94.5\% vs. 93.4\% for NTAA vs. SPOS). There are two reasons behind the improvement of our method over others. First, as explained Section \ref{sect:theory}, our NTAA better accords with the goal and setting of transfer learning. Second, there is a gap between existing NAS solutions and transfer learning. In particular, the target architecture is searched for on the target task (i.e., person re-ID). But the target architecture can not be directly used on the target task until it has been pretrained on the source task (i.e., ImageNet classification).

\begin{table}[t]
\small
\caption{Can the optimal weights of NTAA searched by NAS? (Market-1501) (\textbf{R-1,5,10:} Rank-1,5,10)}\label{tab:nat_retrain}
\centering
\begin{tabular}{c|c|c|c|c}
Method                 & R-1 & R-5 & R-10 & mAP \\ \hline
\multirow{1}{*}{NTAA} & \textbf{{94.5}} & \textbf{{97.9}} & \textbf{{98.3}} & \textbf{{84.8}} \\ \hline
\multirow{1}{*}{Retrain NTAA's architecture} & 92.0 & 96.9 & 97.9 & 77.4 \\
\end{tabular}
\vspace{-11pt}
\end{table}

\begin{figure*}
\centering
\includegraphics[width=0.7 \textwidth]{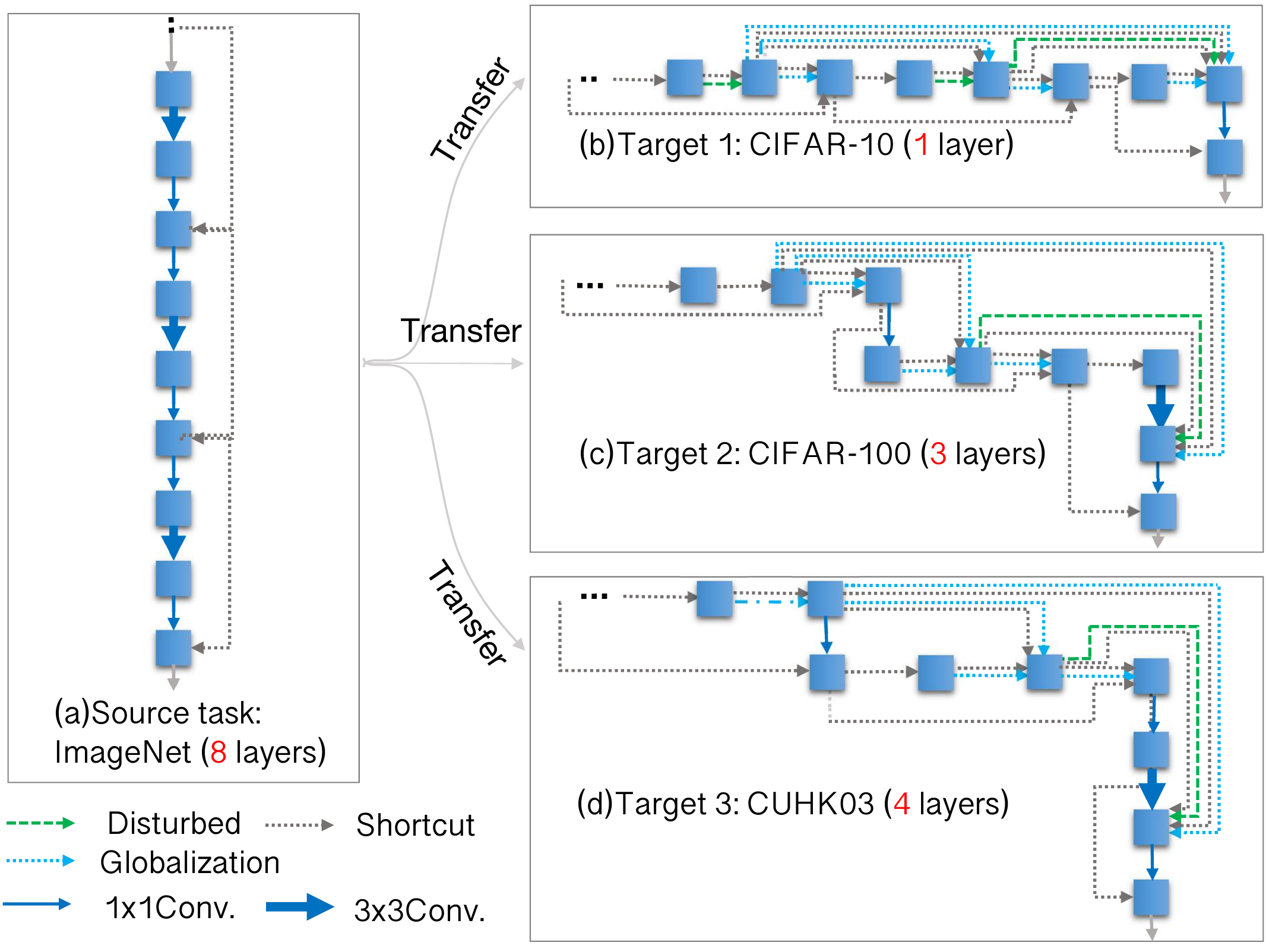}
\caption{{An example of visualization of the neural architecture adaptation from the pre-training on ImageNet to different image recognition tasks (best view in color).}}\label{fig:visual}
\vspace{-11pt}
\end{figure*}

According to the above principled analysis, our NTAA can produce the optimal pair of \{$\alpha_{*}, W_{*}$\} that can not be searched by existing NAS methods. Accordingly, we provide some empirical studies for demonstrating the superiority of our NTAA. We use the optimal optimal pair of \{$\alpha_{*}, W_{*}$\} (i.e., the state-of-the-art model) in Table \ref{tab:market-1501} and Table \ref{tab:nas_nat} as our reference. Specifically, we employ the architecture $\alpha_{*}$ while discarding the network weights $W_{*}$. The network weights are reset as randomly initialized weights. Then we retrain the architecture $\alpha_{*}$ from scratch. The converged accuracy is compared with our NTAA in Table \ref{tab:nat_retrain}. We can observe a significant performance degradation of the retrained model compared with our NTAA. The inaccessibility of $W_{*}$ has also be noticed by \cite{Frankle2019lottery_ICLR,SoelenS2019using_IJCNN,Morcos2019One_CoRR}, proving the importance of our NTAA in synchronously learning the network weights and the architecture in the same way as our NTAA.

\subsection{Image Classification}

We further evaluate our NTAA on the CIFAR-10 \cite{krizhevsky2009learning}, which consists of 10 categories. The models are trained on the 50,000 training images and evaluated on the testing 10,000 images. CIFAR-10 is chosen for the following reasons. First, the data volumes of CIFAR-10 are far less than that of ImageNet. Pretraining on ImageNet may provide performance gains. Second, CIFAR-10 has fewer categories than ImageNet, which consists of 1,000 categories. Learning on more complicated tasks may enable more transferable representations. Finally, the resolution of CIFAR-10 is less than that of ImageNet. Specifically, CIFAR Images are $32\times32$ pixels, while ImageNet images are 500 $\times X$ pixels, where $X$ denotes the width of a rectangle. There is a pattern gap between large images and tiny images, making the finetuning problem more challenging. We examine the error rate. The representative residual networks (e.g., ResNet50 \cite{he2016deep} / ResNet101 \cite{he2016deep} / ResNet152 \cite{he2016deep} / InceptionV2 \cite{szegedy2015going} / VGG \cite{simonyan2014very}) are chosen as our backbones. NTAA is performed by employing the standard protocol \cite{liu2018darts}.

\begin{table*}
\caption{The comparison of NTAA and the standard transfer learning scheme.}\label{tab:baseline}
\centering
\small
\begin{tabular}{c|c|cc|cc|cc|cc}
\multicolumn{2}{c|}{\multirow{2}{*}{Method}}                  & \multicolumn{2}{c|}{ResNet50} & \multicolumn{2}{c|}{ResNet101} & \multicolumn{2}{c|}{InceptionV2} & \multicolumn{2}{c}{VGG19} \\ \cline{3-10}
\multicolumn{2}{c|}{}                             & Error    & Depth    & Error     & Depth    & Error     & Depth     & Error    & Depth   \\ \hline
\multicolumn{2}{c|}{Train from scratch}                    & 6.61     & 50      & 6.53     & 101     & 4.27      & -       & 7.09    & 19     \\ \hline
\multirow{1}{*}{weight pretraining} & full  & 3.90     & 50      & 3.85     & 101     & 3.79      & -       & 6.58    & 19     \\ \cline{2-2}
\multirow{1}{*}{and finetuning}& partial & 3.72     & 50      & 3.71     & 101     & 3.63      & -       & 6.55    & 19     \\ \hline
\multicolumn{2}{c|} {\textbf{NTAA} }     & \textbf{2.43 ($\pm$0.04)}     & \textbf{33}      & \textbf{2.39 ($\pm$ 0.05)}     & \textbf{65}     & \textbf{2.30 ($\pm$0.05)}      & -       & \textbf{6.33 ($\pm$ 0.18)}    & \textbf{13}     \\
\end{tabular}
\vspace{-11pt}
\end{table*}

To demonstrate the effectiveness of our NTAA, we explore two representative standard transfer learning methods, i.e., \textbf{full} standard transfer learning and \textbf{partial} standard transfer learning. Specifically, a network is first pretrained on the ImageNet dataset. Then, for the full standard transfer learning, we tune all the network weights on CIFAR-10; for the partial standard transfer learning, only the network weights in the high-level layers are tuned. Note that the architectures are keep fixed for standard transfer learning, which is different from our NTAA. Besides, another baseline is also introduced, i.e., training from scratch on CIFAR-10 without any pretraining steps. The performances of the three baseline models are presented in Table \ref{tab:baseline}.

We have four observations from Table \ref{tab:baseline} for all the backbones. First, both the full and partial standard transfer learning brings a clear gain when compared with training from scratch (e.g., 3.90\% / 3.72\% vs. 6.61\% for ResNet50). This reflects the importance of pretraining network weights.

Second, the partial standard transfer learning has a slightly better performance than the full standard transfer learning. We believe this observation is of importance because it presents insights into transfer learning. \textbf{1)} the data volume of the target domain is usually smaller than that of the source domain. Consequently, the large network architecture may lead to overfitting. \textbf{2)} The network architecture can be divided into a \emph{general} part and a \emph{specific} part. Loosely, the lower layers may be general, while the high-level layers may be specific. Therefore, fixing the low-level layers while finetuning the high-level layers (i.e., partial standard transfer) leads to an improvement.

Third, our NTAA surpasses both the full and partial standard transfer learning. For example, in ResNet50, our NTAA achieves an error rate of 2.43\%, which is 1.47\% and 1.29\% lower than the full and partial standard transfer learning, respectively. The superiority of our NTAA over the standard transfer learning is attributed to the architecture adaptation. Specifically, in our method, the \emph{general} and \emph{specific} part are adaptively learned, while in partial standard transfer learning, they are handcrafted and fixed.

Fourth, the depth of our NTAA is significantly less than all baseline models (e.g., 33 vs. 50 in ResNet50), indicating that our NTAA has a more efficient neural architecture. The above observations verify the superiority of our NTAA over the standard transfer learning pipelines.

\textbf{Computational Complexity} Both the search \& training process contain 600 epochs. The search takes 1.3 days on 4 GPUs. The training takes 1.9 days on a single GPU.

\subsection{Age Estimation \& Gender Recognition}

To investigate the effectiveness of our NTAA in more complicated tasks, we adopt age estimation and gender recognition tasks on Adience \cite{DBLP:journals/tifs/EidingerEH14}, which is very challenging due to the extreme variations, e.g., low resolution, occlusions, pose and expression variations. Adience consists of $26k$ unconstraint images of 2,284 person IDs, whose ages range from 0 to 60+. Images in the Adience dataset are taken in the wild, making the evaluation on Adience meaningful.

We use the standard evaluation protocol on Adience to measure the top-1 accuracy \cite{DBLP:conf/mipr/LeeCCC18}. Also, following the standard settings, we do not use an additional face alignment technique. In contrast, the official aligned version of faces are used in our experiment. The images are resized to $256\times256$, and a random crop or a center crop of $224\times224$ pixels are applied during the training and testing, respectively.

\begin{table}
\small
\centering
\caption{Age estimation \& gender recognition on Adience.}\label{tab:adience}
\begin{tabular}{c|c|c}
 Method & Age Accuracy & Gender Accuracy \\ \hline
Train from scratch & 53.5\% & 93.8\% \\
pretraining and finetuning & 56.4\% & 94.1\% \\\hline
Levi Hassner \cite{DBLP:conf/cvpr/LeviH15} & 44.1\% & 82.5\% \\
LMTCNN \cite{DBLP:conf/mipr/LeeCCC18} & 44.3\% & 85.2\% \\
WRN \cite{DBLP:conf/bmvc/ZagoruykoK16} & 57.4\%& 93.9\% \\
WRN* \cite{DBLP:journals/corr/abs-1907-13075} & 59.7\% & 94.6\% \\\hline
\textbf{NTAA} & \textbf{64.3\%} & \textbf{95.5\%} \\
\end{tabular}
\vspace{-11pt}
\end{table}

We use the recently proposed EfficientNet-B4 \cite{DBLP:conf/icml/TanL19} as the initial architecture $\alpha_0$. As EfficientNet-B4 \cite{DBLP:conf/icml/TanL19} uses MBConv as its cell, which does not contain a vanilla convolution, we revise our search space $\mathcal{A}$ by replacing the $5\times5$ and $3\times3$ vanilla convolution in the candidate operation sets with $5\times5$ and $3\times3$ separate convolution. We train the EfficientNet-B4 \cite{DBLP:conf/icml/TanL19} for 150 epochs with a cosine learning rate schedule. The batch size is set as 32. Two baselines are adopted, including training from scratch and standard transfer learning. For training from scratch, the initial learning rate is set as 0.1. For standard transfer learning and our NTAA, the initial learning rate is set as 3e-3. Experimental results in Table \ref{tab:adience} shows that our NTAA has a large gain over the two baselines, especially on age estimation (64.3\% vs. 56.4\% for NTAA vs. weight pretraining and finetuning), verifying the effectiveness of our NTAA. We further compare our method with four state-of-the-art methods, including Levi Hassner \cite{DBLP:conf/cvpr/LeviH15}, LMTCNN \cite{DBLP:conf/mipr/LeeCCC18}, WRN \cite{DBLP:conf/bmvc/ZagoruykoK16}, and WRN* \cite{DBLP:journals/corr/abs-1907-13075}. We can see that our NTAA outperforms the competitors by a large margin (e.g., 4.6\% improvement on age estimation). These comparisons demonstrate the effectiveness of our NTAA.

\begin{table}
\small
\centering
\caption{Domain adaptation on DomainNet.}\label{tab:domainnet}
\begin{tabular}{c|c|c|c}
Metric & Models & Accuracy & p-value \\ \hline
Single Best & Source Only & 26.4 ($\pm$0.70) & \multirow{2}{*}{{0.0066}} \\
+ NTAA & Source Only & 28.4 ($\pm$0.68) & \\ \hline
Source Combine & Source Only & 32.9 ($\pm$0.54) & \multirow{2}{*}{{0.0002}} \\
+ NTAA & Source Only & 34.5 ($\pm$0.60) & \\ \hline
Oracle & AlexNet & 58.0 ($\pm$0.53) &\multirow{2}{*}{{0.0079}}\\
+ NTAA & AlexNet & \textbf{{61.1}} ($\pm$0.54) & \\
\end{tabular}
\vspace{-11pt}
\end{table}

\subsection{Unsupervised Domain Adaptation}
We finally validate the effectiveness of our method on the unsupervised domain adaptation task on DomainNet \cite{Peng2019Moment_iccv}, which is currently the largest benchmark for multi-source unsupervised domain adaptation. Specifically, it contains about 0.6 million images belonging to 6 domains and 345 categories. The problem definition of standard transfer learning is clearly different from that of domain adaptation. Standard transfer learning means that a backbone of a model is trained on a large source dataset and is finetuned on a small target dataset. Note that the source task is different from the target task in standard transfer learning. For example, the source task is image classification, while the target task is age recognition. Both the source domain and the target domain have training data and testing data. In contrast, domain adaptation means that a model is trained on the source domain and is tested on the target domain. In domain adaptation, the source task is the same as the target task (e.g., both are 1000-category classification). Moreover, the source domain only contains the training set, and the target domain only contains test set.

The difference between standard transfer learning and domain adaptation makes it infeasible to compare our method and domain adaptation method on DomainNet. Although we cannot compare our method with domain adaptation methods, we can compare our method with the single-source method, single-best method, and the oracle method on Domain. As shown in Table \ref{tab:domainnet}, our method outperforms the baselines remarkably. The present study confirms that our method is promising in closing the domain gap. This indicates a promising direction in searching for architecture in unsupervised domain adaptation tasks, which is our future work.

\subsection{Visualization of Architecture Adaptation.}
To investigate how the same architecture is transferred to different domain tasks, we visualize the target architectures $\alpha^*$ in the three tasks (CIFAR-10, CIFAR-100, and CUHK03), respectively. These tasks share the same initial architecture $\alpha_0$, which is a ResNet50 pretrained on ImageNet. Due to the space limitation, only the 3$^{rd}$ to 10$^{th}$ layers are exhibited in Fig. \ref{fig:visual}, where we have two observations. First, transferred to the smaller dataset or simpler task, the architectures are towards shallower. For example, CIFAR-10 is the simplest task among the above benchmarks, which has only ten categories. Therefore the transferred architecture in CIFAR-10 is shallowest (only \emph{one} layer). On the contrary, the more challenging person re-identification task on CUHK03 with 767 training categories has a deeper transferred architecture (4 layers). Second, in addition to removing the redundant depth, different adaptation tasks prefer different new operations. For instance, the CIFAR-10 task enables more noise disturbing operations than the others. This may attribute to the tiny volume of CIFAR-10 dataset, which requires noise disturbing as data augmentation. The above observations confirm the fact that each domain task with a distinct recognition target may need different levels/paths of feature hierarchy, as our NTAA does.

\subsection{Analysis of Stableness and Robustness}

To validate whether our results are statistically significant, we perform the architecture adaptation for four times and report the means and the error bars. The results are reported in Table \ref{tab:baseline} and \ref{tab:domainnet}. We can find that our method is stably better than the competing methods. For example, using ResNet50 on CIFAR-10, our NTAA outperforms the method of weight pretraining and finetuning by 1.3\%; but the error bar of our method is only 0.04\%. This statistical significance verifies the robustness and effectiveness of our method.

{Apart from the means and error bars, another useful tool to eliminate the randomness in our conclusion is a statistical test. Although no previous work reports a statistical test, a statistical test really does provide new insight into the NAS. It can serve as a new metric to evaluate NAS's effectiveness, which is quite important to the NAS community. To conduct a statistical test, we perform the following five steps. \textbf{Step 1:} State the null hypothesis, i.e., we have $H_0$: There is no significant difference between our method's performance and the competitors. \textbf{Step 2:} State the alternate hypothesis, i.e., we have $H_1$: There is a significant difference between our method's performance and the competitors. \textbf{Step 3:} State the $\alpha$ value. We use the default 0.05 as the $\alpha$ value. \textbf{Step 4:} Perform variance test in the domain adaptation task on DomainNet. Although the mean accuracy of our NTAA is significantly higher than the competitors, we still challenge whether the superiority of our NTAA is caused by randomness. Therefore, we decide to perform a variance test. We run the architecture transfer experiments four times and use the results to calculate the p-values of the variance test \footnote{This can be simply done by using the \emph{anova1} function in MATLAB.}. The statistical test results are reported in Table \ref{tab:domainnet}. As shown, the p-values are 0.0066, 0.0002, and 0.0079. \textbf{Step 5:} Since the p-values are less than the $\alpha$ level 0.05, we have to reject the null hypothesis and accept the alternate hypothesis. This indicates that our NTAA is robust and effective.}

\section{Conclusion}\label{sect:conclusion}

In this paper, we have proposed a novel learning framework called neural transfer and architecture adaptation (i.e., NTAA), which is capable of jointly adapting network weights and network architecture into new domains. The principled analysis has been discussed to explain the reason behind the effectiveness of NTAA by comparing it with the existing solutions to NAS. We have highlighted that preserving the joint distribution of the network architecture and weights is of importance. Our experiments show that the proposed NTAA outperforms state-of-the-art methods on four computer vision tasks while achieving remarkable accuracy improvements.

In future work, a promising direction is to develop a more powerful inference such as the stochastic Monte Carlo algorithms for exploring the optimal neural network architectures.

\section*{Acknowledgement}

This work was supported in part by the National Key Research and Development Program of China under Grant No. 2018YFC0830103, in part by Major Project of Guangzhou Science and Technology of Collaborative Innovation and Industry under Grant 201605122151511, in part by the National Natural Science Foundation of China (NSFC) under Grants 61876045 and 61836012, and in part by Zhujiang Science and Technology NewStar Project of Guangzhou under Grants 201906010057.

%------------------------------------------------------------------------
\bibliographystyle{IEEEtran}
\bibliography{mybibfile}

\begin{IEEEbiography}[{\includegraphics[width=1in,height=1.25in,clip,keepaspectratio]{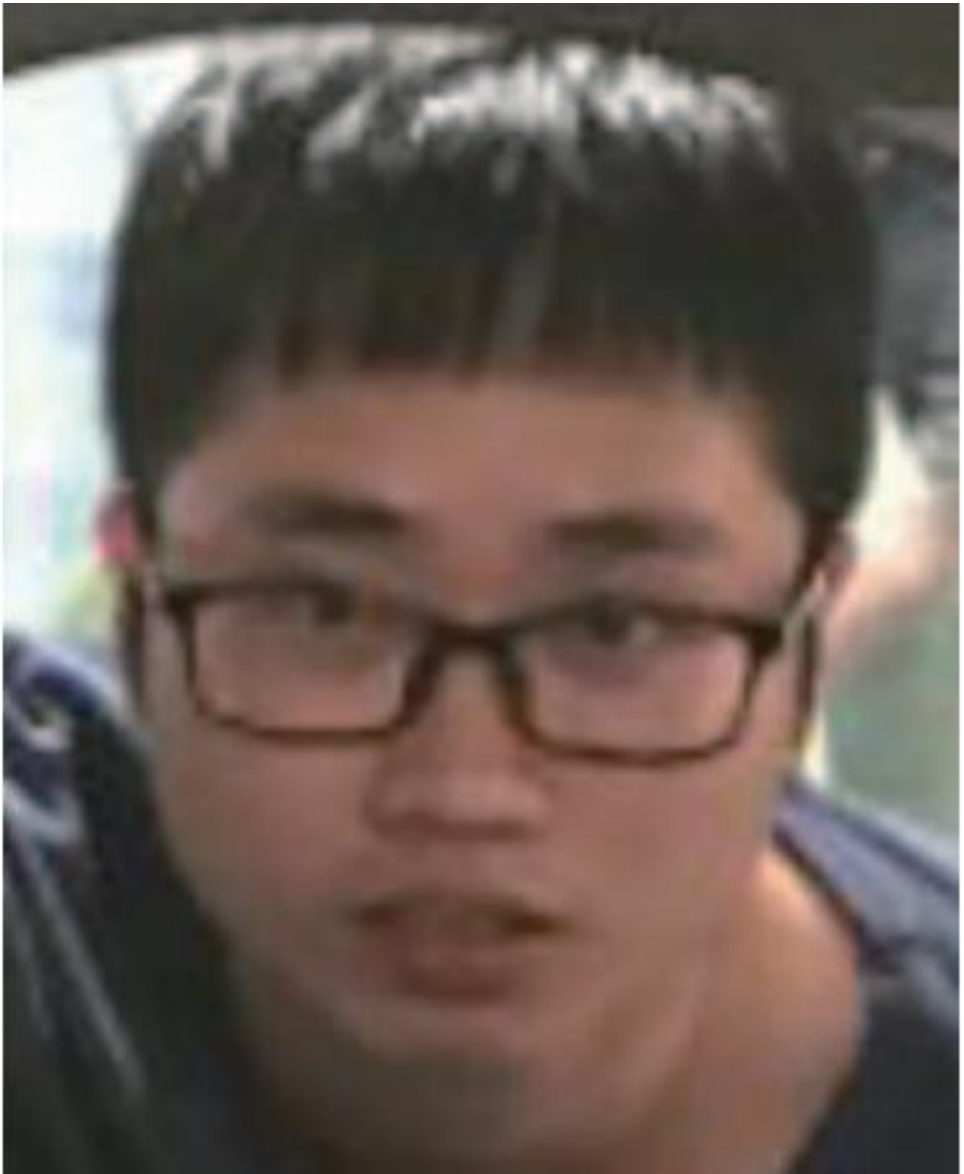}}]{Guangrun Wang} is currently a Ph.D. candidate in the School of Data and Computer Science, Sun Yat-sen University, Guangzhou, China. He received the B.E. degree from Sun Yat-sen University in 2014. From 2015 to 2017, he was a visiting scholar with the Department of Information Engineering, The Chinese University of Hong Kong. His research interests include machine learning and computer vision. He is the recipient of the 2018 Pattern Recognition Best Paper Award and one ESI Highly Cited Paper.
\end{IEEEbiography}

\begin{IEEEbiography}[{\includegraphics[width=1in,height=1.25in,clip,keepaspectratio]{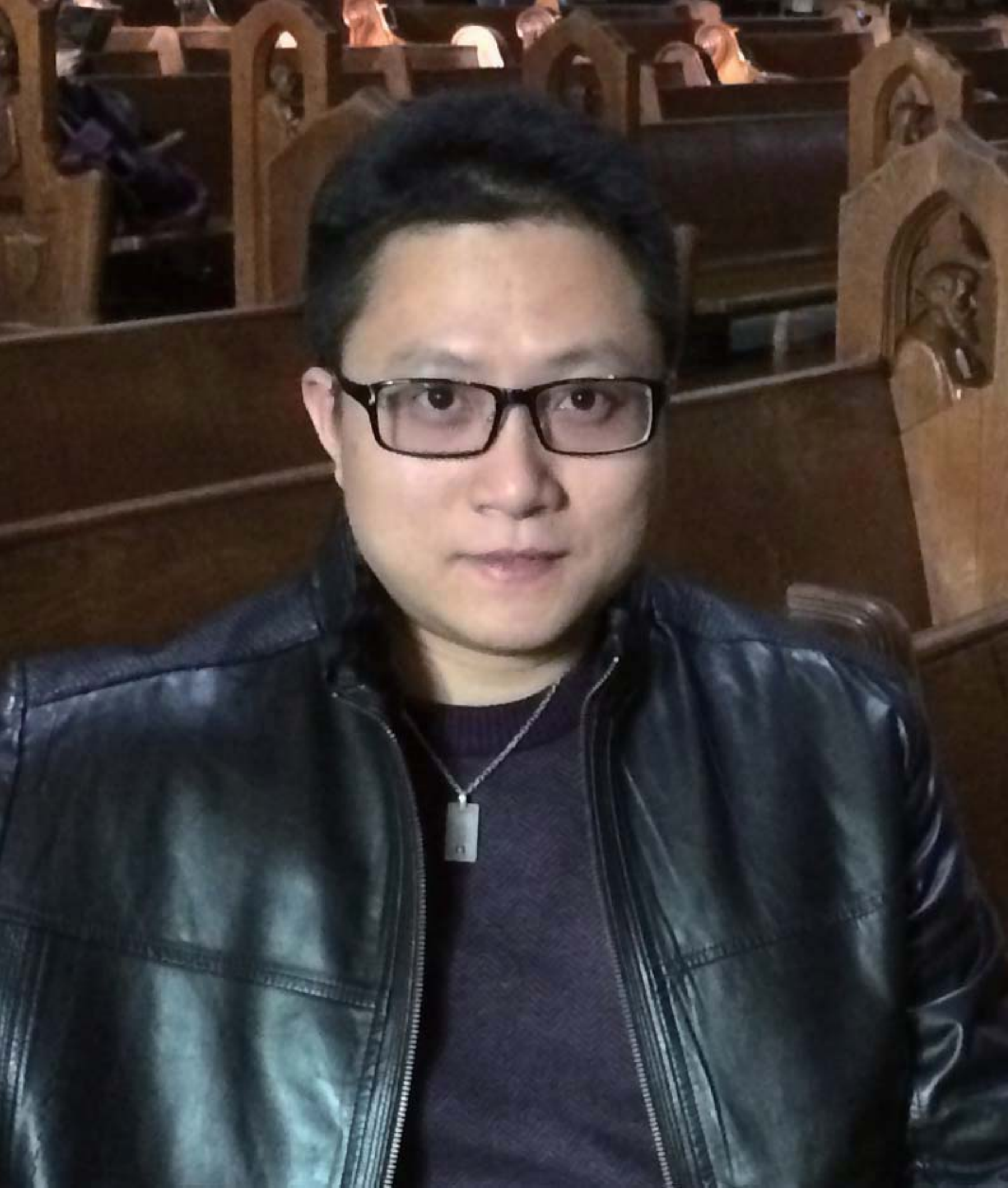}}]{Liang Lin}
(M'09-SM'15) is a Full Professor at Sun Yat-sen University. He served as the Executive R\&D Director and Distinguished Scientist of SenseTime Group from 2016 to 2018, taking charge of cutting-edge technology transfer into products. He has authored or co-authored more than 200 papers in leading academic journals and conferences (e.g., TPAMI/IJCV, CVPR/ICCV/NIPS/ICML/AAAI). He is an associate editor of IEEE Trans, Human-Machine Systems, and IET Computer Vision. He served as Area Chairs for numerous conferences such as CVPR and ICCV. He is the recipient of numerous awards and honors including Wu Wen-Jun Artificial Intelligence Award for Natural Science, ICCV Best Paper Nomination in 2019, Annual Best Paper Award by Pattern Recognition (Elsevier) in 2018, Best Paper Dimond Award in IEEE ICME 2017, Google Faculty Award in 2012, and Hong Kong Scholars Award in 2014. He is a Fellow of IET.
\end{IEEEbiography}

\begin{IEEEbiography}[{\includegraphics[width=1in,height=1.25in,clip,keepaspectratio]{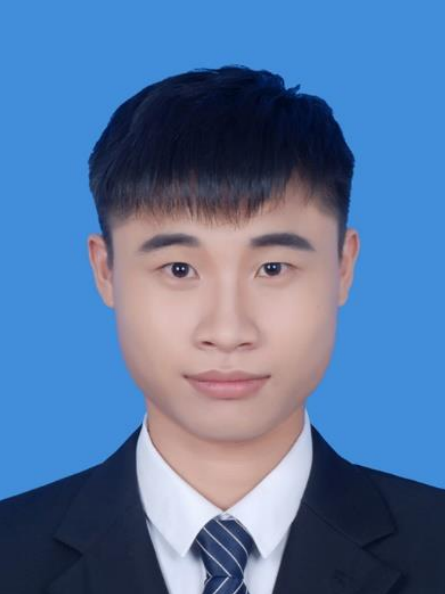}}]{Rongcong Chen} received his B.E. degree from the School of Mathematics, Sun Yat-sen University, Guangzhou, China, in 2017, where he is currently pursuing his Master’s Degree in computer science with the School of Data and Computer Science. His current research interests include computer vision and machine learning.
\end{IEEEbiography}

\begin{IEEEbiography}[{\includegraphics[width=1in,height=1.25in,clip,keepaspectratio]{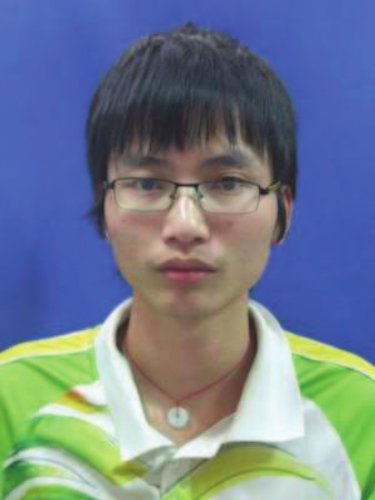}}]{Guangcong Wang} is pursuing a Ph.D. degree in the School of Data and Computer Science, Sun Yat-sen University, Guangzhou, China. He received the B.E. degree in communication engineering from Jilin University (JLU), Changchun, China, in 2015. His research interests are computer vision and machine learning. He has published several works on person re-identification, weakly supervised learning, semi-supervised learning, and deep learning.
\end{IEEEbiography}

\begin{IEEEbiography}[{\includegraphics[width=1in,height=1.25in,clip,keepaspectratio]{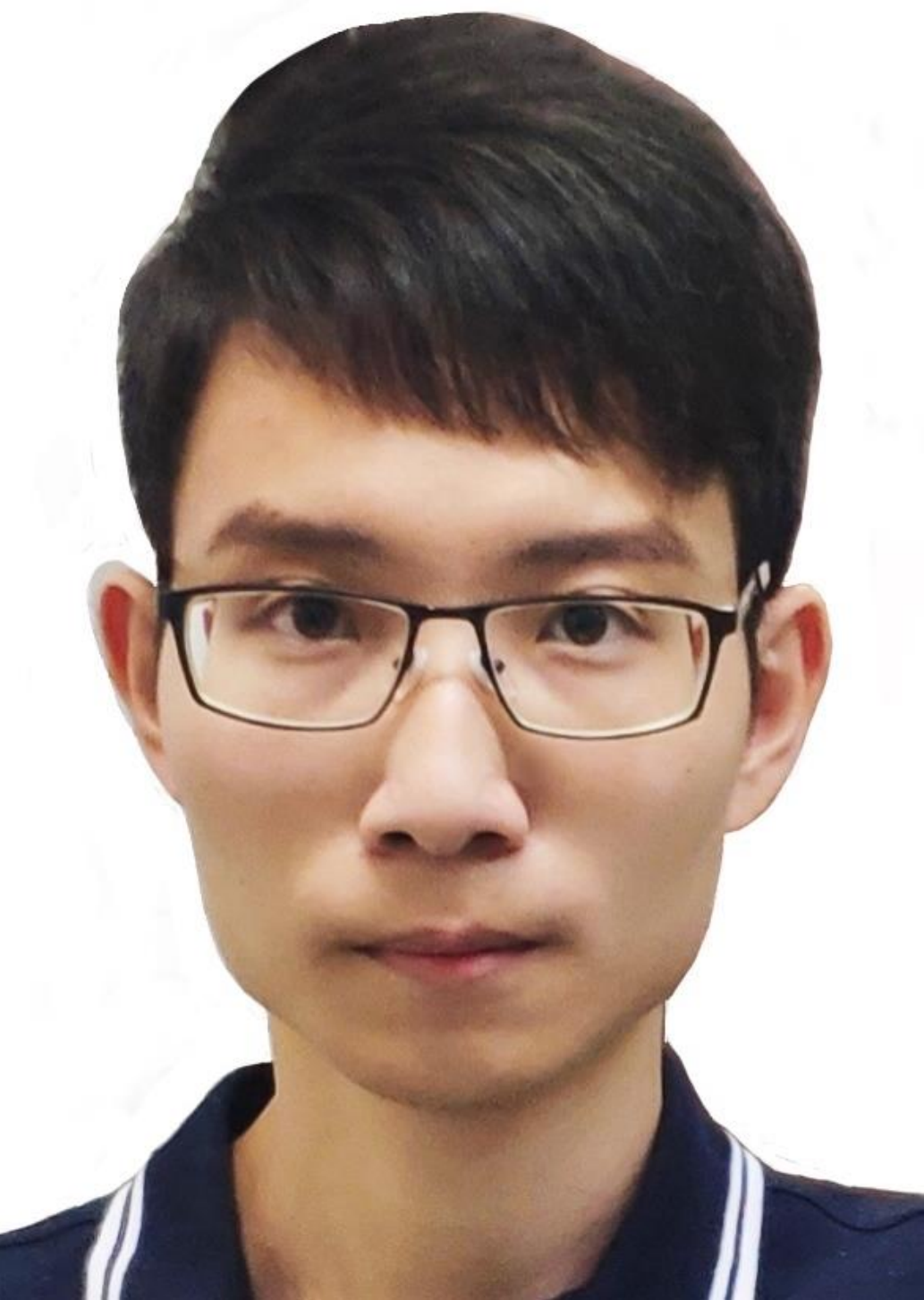}}]{Jiqi Zhang} is currently an undergraduate student in the School of Data and Computer Science, Sun Yat-sen University, Guangzhou, China. His research interests include machine learning and computer vision.
\end{IEEEbiography}

\ifCLASSOPTIONcaptionsoff
 \newpage
\fi
\end{document}